\documentclass[10pt,twocolumn,letterpaper]{article}

\usepackage[pagenumbers]{cvpr} % To force page numbers, e.g. for an arXiv version

\usepackage[dvipdfmx]{graphicx}
\usepackage{caption}
\usepackage{blindtext}
\usepackage{amsmath}
\usepackage{amssymb}
\usepackage[symbol]{footmisc}
\usepackage{booktabs}
\usepackage{mathtools}
\makeatletter
\def\Hline{%
\noalign{\ifnum0=`}\fi\hrule \@height 2pt \futurelet
\reserved@a\@xhline}
\makeatother
\usepackage[pagebackref,breaklinks,colorlinks]{hyperref}

\usepackage{algorithm}
\usepackage{algorithmic}
\usepackage{here}
\usepackage{multirow}

\usepackage[capitalize]{cleveref}
\crefname{section}{Sec.}{Secs.}
\Crefname{section}{Section}{Sections}
\Crefname{table}{Table}{Tables}
\crefname{table}{Tab.}{Tabs.}

\def\cvprPaperID{} % *** Enter the CVPR Paper ID here
\def\confName{AI4CC}
\def\confYear{2023}

\title{Fine-grained Image Editing by Pixel-wise Guidance Using Diffusion Models}
\author{Naoki Matsunaga\footnotemark \qquad  Masato Ishii\qquad Akio Hayakawa\qquad Kenji Suzuki\qquad Takuya Narihira\\
Sony Group Corporation\\
Tokyo, Japan \\
}

\begin{document}
\twocolumn[{%
\renewcommand\twocolumn[1][]{#1}%
\maketitle
\begin{center}
    \centering
    \captionsetup{type=figure}
    \includegraphics[width=0.87\linewidth]{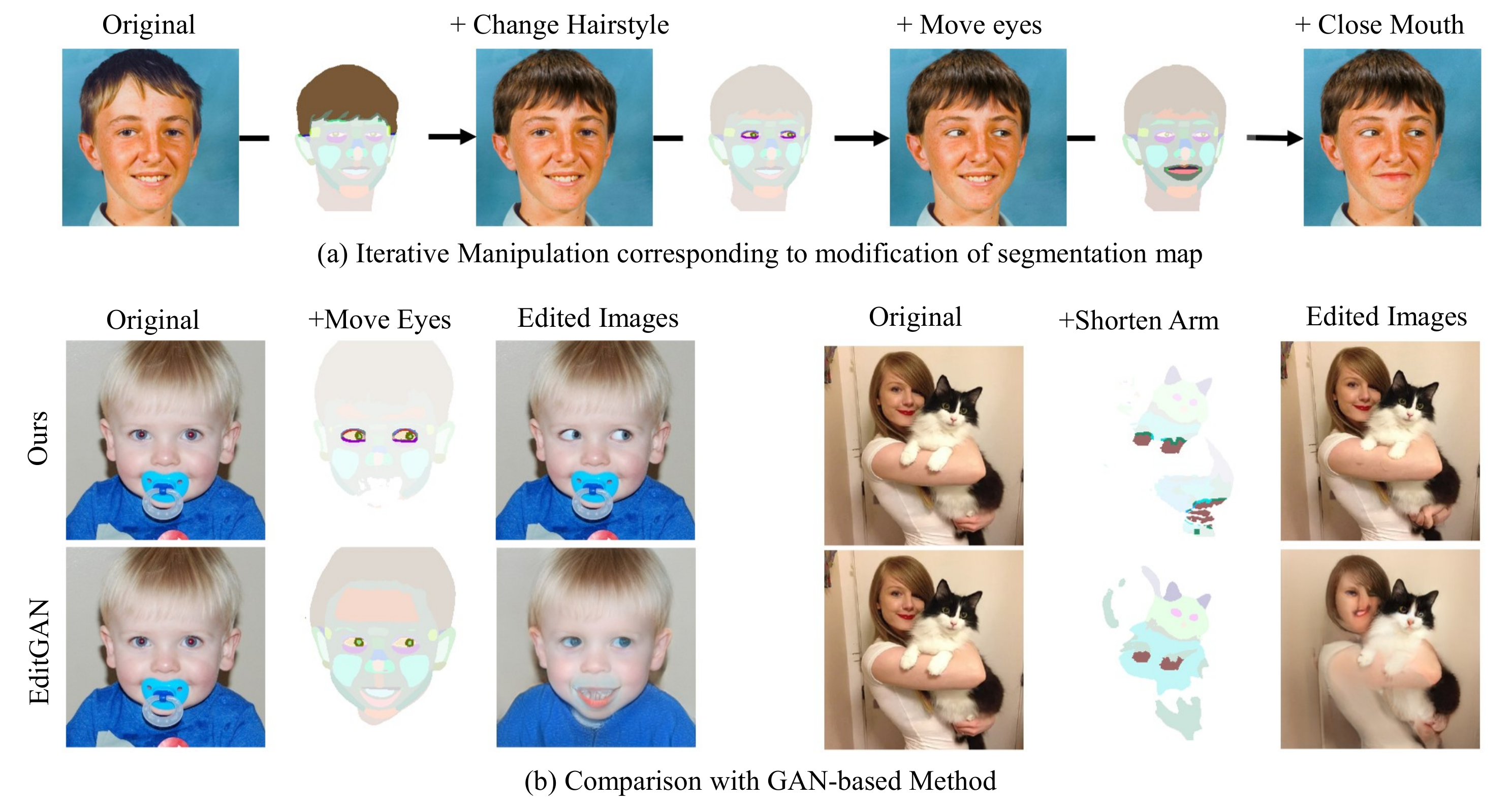}
    \captionof{figure}{Our approach with pixel-wise gradient-based guidance enables fine-grained real-image editing. An existing GAN-based method, called EditGAN, fails to reconstruct detailed features, which are less observed in the training data (\textit{e.g.,} a pacifier in the left case). In contrast, our diffusion-based method enables pixel-wise editing while preserving detailed features.}
\label{fig:overall_results}
\end{center}%
}]

%--------------------------------------------------------------------

%%remove sideline number
% https://stackoverflow.com/questions/71927123/can-anyone-tell-me-how-can-i-remove-this-side-line-number-and-all-blue-color-wri

%% Abstract
\begin{abstract}
Our goal is to develop fine-grained real-image editing methods suitable for real-world applications.
In this paper, we first summarize four requirements for these methods and propose a novel diffusion-based image editing framework with pixel-wise guidance that satisfies these requirements.
Specifically, we train pixel-classifiers with a few annotated data and then infer the segmentation map of a target image.
Users then manipulate the map to instruct how the image will be edited.
We utilize a pre-trained diffusion model to generate edited images aligned with the user's intention with pixel-wise guidance.
The effective combination of proposed guidance and other techniques enables highly controllable editing with preserving the outside of the edited area, which results in meeting our requirements.
The experimental results demonstrate that our proposal outperforms the GAN-based method for editing quality and speed.
Code is available at \url{https://github.com/sony/pixel-guided-diffusion.git}~\footnote[2]{Code is implemented by NNabla~\cite{narihira2021neural} and pytorch~\cite{paszke2019pytorch}.} 
\end{abstract}
\footnotetext[1]{Corresponding author \href{mailto: Naoki.Matsunaga@sony.com}{Naoki.Matsunaga@sony.com}}

%% Introduction
% \section{Introduction}
% \label{sec:intro}
% \input{subtex_ws/introduction}
% \vspace{-8mm}
% Introduction
\section{Introduction}
\label{sec:intro}
%% 問題設定とその難しさを述べる
% 近年生成モデルは様々な画像編集に利用されており、実用化も進んでいる。
% ユーザーの意図を画像の細部まで直感的に反映させる方法として、本研究では、Fine-grained Editingに着目する。
% 本論文では、Fine grained Editingとは、画像内に物体を生成するのではなく、元のオブジェクトのAppearanceなどを局所的に編集するピクセルレベルの編集手法のことをさす
% これらの手法を実応用する際には以下の要件：
% (1) 実画像に対して編集可能であること(編集領域外を保持したまま、ピクセル単位の編集が可能)
% (2) 編集領域外を保持しつつ編集領域内の編集をユーザーの意図に沿って実現する
% (3) 高速な編集が可能である。
% を同時に満たすことが必要であり、これらを同時に満たすことで、leads to higher controllability and better user experience　in the editing process.
Recently, generative models have been widely utilized in many image-editing methods ~\cite{Croitoru2022DiffusionMI, Xia2022GANIA} thanks to their potential to generate high-quality visual content. 
Several techniques have already been applied to real-world applications by digital artists~\cite{bailey2020tools} and have contributed to expanding their creativity. 
To reflect the user's intention to the details in edited images more effectively, we focus on fine-grained real image editing.

We refer to fine-grained image editing as an editing process with pixel-level control that does not generate new content but makes small and localized changes to the appearance of existing content in the image, such as manipulating facial parts and animal pose shown in Fig.\ref{fig:overall_results}.
Considering the real-world application of this editing process, it requires four properties:
\begin{description}
\setlength{\parskip}{0cm}
\setlength{\itemsep}{0cm}
\item[P.1] visual content in the editing area should be created aligned with the user's intent down to the pixel,
\item[P.2] the content outside the edited area should be precisely preserved, and
\item[P.3] it can be built in a label-efficient manner to apply it to a new category easily, and
\item[P.4] fast editing should be possible.
% \vspace{-2mm}
\end{description}

% \hspace{-3mm}
Jointly meeting these requirements leads to higher controllability and better user experience.

%% 既存研究となぜ解けないのかを述べる
In the literature, GANs (Generative Adversarial Networks) ~\cite{goodfellowgenerative} have been intensely adopted. 
GAN-based methods can precisely reflect fine-grained instruction by users, such as manipulated semantic segmentation map ~\cite{Lee2020MaskGANTD, Park2019SemanticIS,Zhu2020InDomainGI,Ling2021EditGANHS} , skethes~\cite{Chen2020DeepFaceDrawingDG} into edited images. 
Notably, EditGAN~\cite{Ling2021EditGANHS} is specifically designed to learn such editing in a label-efficient manner.
Although these methods successfully provide high-quality visual content in the edited area, it is essentially difficult to strictly preserve other contents in the target image (\textit{i.e.} P.2), as shown at the bottom of Fig.\ref{fig:overall_results} because it requires solving a challenging problem which is to find the latent features corresponding to the edited image while satisfying such constraints.
In summary, many GAN-based methods satisfy P.1 and P.4, and EditGAN further satisfies P.3, 
but it is inherently difficult for these methods to satisfy P.2, which would result in a serious barrier to their use in real-world applications.
% It could be a significant drawback of real image editing applications.

%% 提案手法とその有効性や貢献について述べる
% 上記の課題を解決し4つの要件を満たすためにDiffusion Modelを利用した新規画像操作手法を提案する。
% 提案手法では、ラベル効率のよいSegmentation ModelであるDatasetDDPMをGuidance Modelとして利用し、ピクセル単位のGuidance Lossを提案し、Lossを利用してReverse Processをガイダンスるする。
% ガイダンス時には、lossとblended diffusionをうまく組み合わせることで、オリジナルの実画像のコンテンツを保持した編集を実現する。
% また、ピクセル単位のRespaced Guidanceを実験し、GAN based の手法に匹敵する編集時間を実現する。
% 実験においては、GAN-basedの手法に対して、定量的、定性的に性能向上することを確認した。

To address the above-mentioned issue while meeting four properties, 
we propose a novel diffusion-based editing method that enables fine-grained editing by user-provided manipulated segmentation map as shown in Fig.~\ref{fig:overall_results}. 
Compared to the prior works, we use diffusion models instead of GANs. It is effective to satisfy P.2, but makes it challenging to satisfy P.3 and P.4 in general.
Therefore, we extend DatasetDDPM~\cite{Baranchuk2022LabelEfficientSS}, which is a label-efficient semantic segmentation model, 
to every time step and use it to guide the generation process of reverse diffusion process based on pixel-wise segmentation loss (for  P.1 and P.3).
Additionally, we adopt accelerated generation ~\cite{Dhariwal2021DiffusionMB} to this pixel-level guidance to achieve comparable editing speed to the GAN-based method (for P.4).
Experimental results validate the advantage of our method both quantitatively and qualitatively.

In summary, our contributions are three-fold: 
\begin{itemize}
\setlength{\parskip}{0cm}
\setlength{\itemsep}{0cm}
\item We reveal that fine-grained real image editing  for real-world applications can be achieved with an effective combination of pixel-wise guidance and other techniques.
\item We reveal that the optimal setting of both start denoising step $t_0$ and guidance scale $s$ depends on the size of the edited area. This is helpful in determining the parameter setting for unseen data on applications.
% % \item We reveal that there are disentangled latent vectors on the latent space of the pre-trained ADM. It is helpful to control the intensity of editing.
\item Our pixel-wise guidance is more computationally efficient than the conventional guidance using full-size segmentation models. It contributes to fast editing.
% % \item Our method achieved computationally more efficient guidance than using full-size segmentation models for guidance. It is helpful for fast editing.

\end{itemize}

%% Related Works
\section{Related Works}
\label{sec:related}
\subsection{Diffusion models}
\label{subsec:diffusion models}
Denoising Diffusion Probabilistic Model (DDPM)~\cite{Ho2020DenoisingDP,SohlDickstein2015DeepUL,Nichol2021ImprovedDD,Dhariwal2021DiffusionMB} is a generative model that aims to generate data by reversing a diffusion process.
The forward diffusion process is defined as a Markov chain process with time, where a small amount of Gaussian noise is added to samples in $T$ steps, resulting in $x_T$ following an isotropic Gaussian distribution.
A single step in this process can be represented by the following equation: $q(x_t | x_{t-1}) \coloneqq \mathcal{N}(\sqrt{1-\beta_t}x_{t-1}, \beta_t I)$, where $\beta_t$ is the hyperparameter that defines the process.
If $\beta_t$ is sufficiently small, the step of the reverse diffusion process  $q(x_{t-1}| x_t)$ can also be parameterized by a Gaussian transition as follows: $p_\theta(x_{t-1}|x_t) \coloneqq \mathcal{N}( \mu_\theta(x_t,t), \Sigma_\theta(x_t,t)),$
the mean $\mu_\theta$ and covariance $\Sigma_\theta$  is estimated by $x_t$ which is an input of the noise estimator $\epsilon_\theta(x_t,t)$, 
where $\theta$ is parameterized by the U-net encoder and the decoder.
The model is trained to obtain $\mu$ via noise estimation for $x_t$, while $\Sigma$ is assumed to be fixed in DDPM.
Consequently, we can represent a single step in the reverse process as follows:
% Denoising Diffusion Probabilistic Model (DDPM) ~\cite{Ho2020DenoisingDP,SohlDickstein2015DeepUL,Nichol2021ImprovedDD,Dhariwal2021DiffusionMB} is a type of latent variable models which has two processes, forward and reverse diffusion process.
% The forward process is a Markov chain process where is adding the noise to the data, obtaining the latent variables $x_t$ for $t=1,... T$. One step of this process can be defined as a gaussian transition $q(x_t | x_{t-1}) \coloneqq \mathcal{N}(\sqrt{1-\beta_t}x_{t-1}, \beta_t I)$.
% Finally, DDPM approximates raw images distribution by the endpoint of this process, which is typically isotropic gaussian distribution.
% In the reverse process, DDPM approximates the added noise by noise estimating model $\epsilon_\theta(x_t,t)$.
% The reverse process $q(x_{t-1}| x_t)$ also can be parameterized by gaussian transition $p_\theta(x_{t-1}|x_t) \coloneqq \mathcal{N}(x_{t-1}; \mu_\theta(x_t,t), \sigma_\theta(x_t,t) I)$. 
% Then, we approximate $x_{t-1}$ as follows:
%%---------------------------
\begin{equation}
\label{eq:ddpm_reverse}
    x_{t-1} = \frac{1}{\sqrt{1-\beta_t}}\Biggl(x_t - \frac{\beta_t}{\sqrt{1-\bar{\alpha}_t}}\epsilon_\theta(x_t,t)\Biggr) + \sigma_t \epsilon,
\end{equation}
%%---------------------------
where $\alpha_t:=1-\beta_t, \bar{\alpha}_t:=\prod_{s=1}^t \alpha_s, \epsilon \sim \mathcal{N}(0,I)$.
Given an initial $x_T$, we generate a sample by repeating the above step from $t=T$ to $t=0$.

Data generation through DDPM is stochastic, as shown in Eq.~\ref{eq:ddpm_reverse}. Therefore, given a fixed $x_T$, we can obtain various $x_0$ through this process. 
Conversely, \cite{Song2021DenoisingDI} extends DDPM to obtain a deterministic generation process by utilizing a non-Markovian forward process.
A single step of the reverse process in DDIM is described as follows:
% This probabilistic sampling process (eq.\ref{eq:ddpm_reverse}) allows for the generation of diverse images. On the other hand, \cite{Song2021DenoisingDI} proposed a non-Markovian noising process as bellow: 
%%---------------------------
\begin{align}
\label{eq:ddim_reverse}
    x_{t-1} = &\sqrt{\bar{\alpha}_{t-1}}f_\theta(x_t, t) + \sqrt{1-\bar{\alpha}_{t-1}}\epsilon_\theta(x_t,t),
\end{align}
%%---------------------------
\begin{equation}
\label{eq:ddim_forward_f}
    f_\theta(x_t, t) \coloneqq \frac{x_t- \sqrt{1-\bar{\alpha}_t}\epsilon_\theta(x_t,t)}{\sqrt{\bar{\alpha}_t}}.
\end{equation}
%%---------------------------

% In eq.\ref{eq:ddim_reverse}, if $\sigma_t = 0$, the sampling process of DDIM would be deterministic.
In this setting, given target data $x_0$, we can obtain the latent variable $x_T$ to generate $x_0$ through DDIM by repeating the following computation: 
% We can obtain the latent variable considering inverted process of this as eq.\ref{eq:ddim_forward}.
%%---------------------------
\begin{equation}
\label{eq:ddim_forward}
    x_{t+1} = \sqrt{\bar{\alpha}_{t+1}} f_\theta(x_t, t) + \sqrt{1-\bar{\alpha}_{t+1}} \epsilon_\theta (x_t, t).
\end{equation}
%%---------------------------
Using Eq.~\ref{eq:ddim_forward} and Eq.~\ref{eq:ddim_reverse}, DDIM achieves accurate reconstruction (inversion).

%%----------------------------------------------------------
\paragraph{Classifier Guidance}
\label{subsubsec:classifier_guidance}
Classifier guidance ~\cite{Dhariwal2021DiffusionMB} is a technique that enables pre-trained unconditional diffusion models to generate images corresponding to a specified class label.
It requires a pre-trained classifier to specify the desired class and uses it to guide the generation process of the diffusion models.
Specifically, classifier $p_\phi (y|x_t, t)$ is first trained with noisy images $x_t$ that correspond to each time step $t$ in the diffusion process.
Then, in the generation process, the mean of the distribution of $x_t$ is shifted at each time step by adding the gradients of the classifier with respect to the data, as follows:
\begin{align}
\label{eq:classifier_guidance}
    p_\theta(x_{t-1}|x_t) =\mathcal{N}(\mu_\theta(x_t,t) + s \Sigma \nabla_{x_t} \mathrm{log}\ p_\phi(y | x_t, t), \Sigma),
\end{align}
where $s$ is the guidance scale parameter. 
% When we set a large value to $s$, the generated images are highly aligned to the desired class, but they are likely to lose their diversity.
% Classifier guidance ~\cite{Dhariwal2021DiffusionMB} is one of the techniques for pretrained unconditional diffusion models to generate specific class-label images by conditioning the sampling process using the gradient of a classifier.
% In this method, we have to train the classifier $p_\phi (y|x_t, t)$ on a noise image $x_t$ which is corresponding to each time-step $t$ in advance.
% Then, guide the diffusion sampling process in the direction of generating the desired class $y$ according to the gradients $\nabla_{x_t} log p_\phi(y | x_t, t)$ as follows:

\paragraph{Semantic Segmentation with Diffusion Models}
\label{subsubsec:semantic segmentaton with diffusion models}
The proposed method builds on DatasetDDPM~\cite{Baranchuk2022LabelEfficientSS}, which is a label-efficient semantic segmentation model that uses pixel-level representations of the noise estimator $\epsilon_\theta$.
The authors called the model trained on synthetic images is called DatasetDDPM. Therefore, this paper calls the model trained on real images as DatasetDDPM-R.
% DatasetDDPM-R estimates a segmentation label on each pixel via a three-layer multi-layer perceptron (MLP), which is referred to as a pixel-classifier, using pixel-level representations, the UNet’s intermediate activations are upsampled and concatenated as inputs.
DatasetDDPM-R estimates a segmentation label on each pixel via a three-layer multi-layer perceptron (MLP), which is referred to as a pixel-classifier. The input of this classifier is the UNet’s intermediate activations which are upsampled and concatenated.
% Because pixel-wise features can be used as one piece of data, it can build a model with few annotated data.
Furthermore, DatasetDDPM~\cite{Baranchuk2022LabelEfficientSS} outperforms DatasetGAN~\cite{Zhang2021DatasetGANEL} in segmentation estimation performance because real images can be used for training.

%%---------------------------------------------------------
\subsection{Image Editing Methods}
\paragraph{GAN-based methods}
\label{subsubsec:gan-based methods}
Generative Adversarial Network (GAN) ~\cite{goodfellowgenerative}-based editing methods are divided into various categories.
There are those, for example, that find a meaningful and disentangled direction in latent spaces in an unsupervised manner~\cite{Xu2022TransEditorTD, Shen2020InterpretingTL, Shen2022InterFaceGANIT, Bau2019GANDV, Bau2019SemanticPM,Plumerault2020ControllingGM, Wu2021StyleSpaceAD} and those that use  text~\cite{Kwon2022CLIPstylerIS,Patashnik2021StyleCLIPTM, Gal2021StyleGANNADACD, Xia2021TediGANTD}.
These methods are helpful for concept-level editings, such as style transfer and attribute transformation, but are unsuitable for localized and fine-grained editing.

To achieve fine-grained editing, some approaches use segmentation maps and sketches for user-provided inputs~\cite{Lee2020MaskGANTD, Park2019SemanticIS,Zhu2020InDomainGI,Chen2020DeepFaceDrawingDG}.
These methods achieve precise editing of details corresponding to the manipulation of user-provided inputs,
but their drawback is that a large amount of annotated data is required for conditional training.
EditGAN~\cite{Ling2021EditGANHS} solve this problem by building on DatasetGAN~\cite{Zhang2021DatasetGANEL}, which consists of a simple three-layer MLP classifiers with the middle features of StyleGAN~\cite{Karras2019ASG, Karras2020AnalyzingAI} as inputs.
This classifier predicts the segmentation label on each pixel, and as a result, EditGAN can achieve high-precision editing in a label-efficient manner.

However, when editing real images, the GAN-based methods listed above first need to search for appropriate latent variables that can reconstruct the images via a generator.
Although, there are many inversion techniques~\cite{Abdal2019Image2StyleGANHT,Abdal2020Image2StyleGANHT,Alaluf2021ReStyleAR,Richardson2021EncodingIS,Tov2021DesigningAE,Zhu2020InDomainGI, Alaluf2022HyperStyleSI} for searching such latent variables, 
these GAN-based methods, including EditGAN suffer from the incomplete reconstruction of features that are scarce in training data.
That could be a major drawback limiting their real-world editing applications.

\paragraph{Diffusion-based methods}
\label{subsubsec:diffusion-based methods}
Diffusion models ~\cite{Ho2020DenoisingDP,SohlDickstein2015DeepUL,Nichol2021ImprovedDD,Dhariwal2021DiffusionMB} and score-based generative models~\cite{Song2019GenerativeMB, Song2021ScoreBasedGM} are suitable for real image editing because it has outperformed image synthesis quality ~\cite{Dhariwal2021DiffusionMB} against existing generative models~\cite{brock2018large, kingma2013auto, van2016pixel, menick2018generating, rezende2015variational} and can be accurately reconstructed, as described in the Sec.\ref{subsec:diffusion models}.
Some works have been conducted in a global manner~\cite{Kim2022DiffusionCLIPTD, Preechakul2022DiffusionAT}, 
which can be applied to style and attribute transformations,
but is unsuitable for localized, fine-grained editing.

For fine-grained editing, some works can be divided into two categories.
One line of research is training the conditional diffusion models~\cite{Nichol2022GLIDETP, Saharia2022PaletteID}.
We can edit images locally using an inpainting technique ~\cite{Saharia2022PaletteID} and text-prompts~\cite{Nichol2022GLIDETP} by training diffusion models using such conditions as inputs.
However, these methods incur high training costs because they are required to train the diffusion models.

On the other hand, unconditional model based methods~\cite{Choi2021ILVRCM,Meng2021SDEditIS,avrahami2022blended,Avrahami2022BlendedLD}, it can use the pre-trained diffusion models, achieve editing with low training cost.
ILVR~\cite{Choi2021ILVRCM} is an image translation method in which the low-frequency component is used for conditioning the sampling process, and it achieves scribble-based editing.
In SDEdit~\cite{Meng2021SDEditIS}, user-provided strokes on images are first noised in the latent spaces by a stochastic SDE process.
The edited images are then generated by denoising via the reverse SDE process.
These methods achieve user-provided localized editing, but they cannot easily edit  fine-grained features. 
In particular, for SDEdit, 
there is no guarantee that the edited content will be preserved through the generation process because no conditions related to the edit are given in a simple stochastic reverse SDE process.
Blended-diffusion ~\cite{Avrahami2022BlendedLD, avrahami2022blended} is a technique for localized editing that corresponds to user-provided text prompts. 
But it is unsuitable for fine-grained editing, such as "Growing hair a little longer," via the text descriptions.

Therefore, we propose the localized and fine-grained image editing method in a label-efficient manner.

\section{Fine-grained Edit with Pixel-wise Guidance}
\label{sec:proposed}
\begin{figure*}[t]
    \centering
    \includegraphics[width=0.91\linewidth]{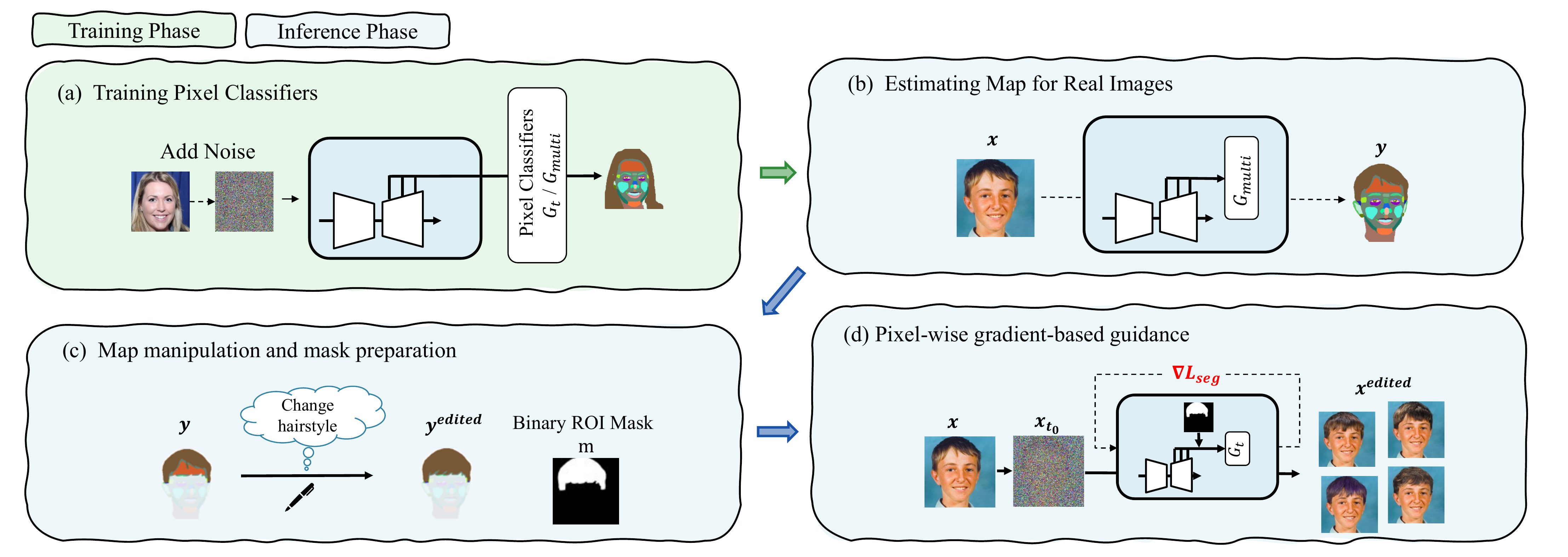}
    \caption{Overall workflow of the proposed method. (a) For preparation, we train pixel classifiers of DatasetDDPM-R~\cite{Baranchuk2022LabelEfficientSS}, which is an MLP-based segmentation model, and save parameters. (b) After training, we estimate the segmentation map using intermediate representations across multiple steps (50,150,250 steps concatenate). (c) We then manipulate the segmentation map into desired one. (d) In the iterative reverse denoising process, we guide the sampling process with the gradient of a pixel classifier $G_t$.}
    \label{fig:method/overall}
\end{figure*}

\subsection{Overview}
\label{subsec:methods/overview}
The overall flow of the proposed method for image editing is shown in Fig.~\ref{fig:method/overall}.
First, we train pixel classifier ~\cite{Baranchuk2022LabelEfficientSS} $G_{multi}$ and $G_t$, which is used to estimate a segmentation map for editing and pixel-wise guidance, respectively.
Then, we denote the set of $G_t$ by $G$.
For map manipulation, we generate the original segmentation map $y$ of a target real image $x$ by applying the pixel classifier $G_{multi}$ to $x$ and then manipulate it to the desired map $y^{edited}$.
Furthermore, we guide the reverse diffusion process with the gradient of a pixel classifier $G_t$ with respect to the data at each time step $t$.
Finally, we obtain edited images $x^{edited}$.
Detailed descriptions are given in the following sections.

% \subsection{Pixel-wise semantic segmentation guidance}
\subsection{Editing Framework}
\label{subsec:methods/main}
\paragraph{Training Pixel Classifiers}
\label{subsubsec:training}
We use the pixel classifiers $G$ of DatasetDDPM-R for gradient-based guidance. This is because we can use them to
(a) build segmentation models in a label-efficient manner and,
(b) reduce the computational cost because noise and label estimation can be computed simultaneously on the same forward path; furthermore, we can
(c) apply it to any pre-trained non-conditional diffusion model.

Similar to the classifier guidance mentioned in Sec.\ref{subsubsec:classifier_guidance}, the intuitive approach is to extend the classifier to accept time step $t$ as input, but this is difficult with tiny MLP-based classifier, so in this study, $G_t$ is trained at each time steps $t$.
When training a pixel-classifier $G_t$, we use image-label pairs, which require human annotation.
For each time step $t$, we use noise samples $x_t$, obtained directly from data $x_0$, as training data.
%%------------------------------
\begin{align}
\label{eq:DatasetDDPM/Train}
    x_t = \sqrt{\bar{\alpha_t}}x_0 + \sqrt{1-\bar{\alpha_t}}\epsilon, \quad \epsilon \sim \mathcal{N}(0,I).
\end{align}
%%------------------------------
Here, the label annotated on $x_0$ is associated with $x_t$ and used for training as image-label pairs.
Then, we save the parameters of $G_t$.
As we mentioned on Sec.\ref{subsubsec:semantic segmentaton with diffusion models}, $G_t$ estimates the segmentation label on each pixel, and the intermediate activations are extracted from the U-Net decoder.

For estimating the original map $y$ as described in Fig.~\ref{fig:method/overall} (b), we train another pixel-classifier $G_{multi}$ with the representations across multiple time steps, which are extracted from $t$=\{50, 150, 250\} and concatenated, following the original paper~\cite{Baranchuk2022LabelEfficientSS}.

\paragraph{Map manipulation and mask preparation}
\label{subsubsec:map manipulation and mask preparation}
After estimating the original segmentation map $y$ using $G_{multi}$ , 
the user can manipulate map $y$ and then obtain the target map $y^{edited}$ as described in Fig.~\ref{fig:method/overall} (c).
Simultaneously, we set the binary ROI mask $m$ by picking up all edit-related pixels from the original map $y$ and manipulated map $y^{edited}$ as follows~\cite{Ling2021EditGANHS}.
%%------------------------------
\begin{equation}
    % m=\left\{p: c_p^{\mathrm{y}} \in Q_{edit}\right\} \cup\left\{p: c_p^{\mathbf{y}^{edited}} \in Q_{edit}\right\}
    m_{ij} = 
    \begin{cases}
    1 & \mathrm{if}\ y_{ij} \in Q_{edit}\ \mathrm{or}\ y^{edited}_{ij} \in Q_{edit}, \\
    0 & \mathrm{otherwise.}
    \end{cases}
\label{eq:roi}
\end{equation}
%%------------------------------
$m\in \mathbb{R}^{H \times W}$ is a binary mask, defined by all pixels $p$ whose segmentation labels $y_p$ are included in the edit-related class list $Q_{edit}$, which is to be preliminarily specified manually.
To seamlessly blend the edited region with the outside of the region, this $m$ is dilated for three pixels as a buffer.

%%------------------------------
\begin{figure*}[t]
    \centering
    \includegraphics[width=0.7\linewidth]{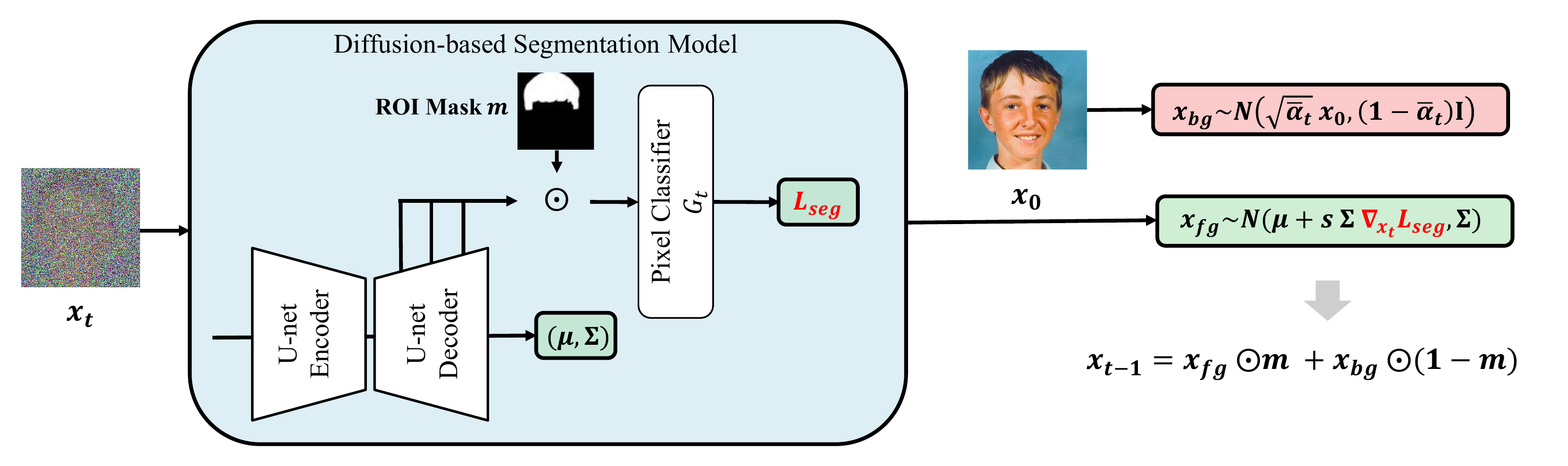}
    \caption{Overview of pixel-wise segmentation guidance at a time-step $t$. The input of the segmentation model is the masked region's ($m$) pixel features. We obtain mean and variance on the normal denoising process and gradient of segmentation loss on the segmentation model's output. Then we can obtain guided foreground images that are in the ROI by pixel-wise guidance.}
    \label{fig:proposed_method}
\end{figure*}
%%------------------------------

\paragraph{Pixel-wise gradient-based guidance}
\label{subsubsec:pixel wise guidance}
To generate images corresponding to the edited map $y^{edited}$, we guide the reverse diffusion process using this map.
First, we obtain the starting noise sample $x_{t_0}$ that was determined by $x_0$ using Eq.\ref{eq:ddim_forward}.
Then, we guide the sampling process at each time step $t$ for $t = t_0, ... 1$, and an  overview of the pixel-wise guidance at each time step is shown in Fig.~\ref{fig:proposed_method}.

% At each time step $t$, the mean and covariance of $x_{t-1}$ was estimated by $x_t$ which is an input of noise estimator $\epsilon_{\theta}(x_t,t)$, parameterized by the U-net encoder and decoder.
Along with noise estimation, we extracted the intermediate activations $z \in \mathbb{R}^{H \times W \times d}$, which are appropriately upsampled and concatenated. 
Following ~\cite{Baranchuk2022LabelEfficientSS}, we set $d$=2816, which is the dimension of intermediate representations at a single time step, in this paper.
For this guide, we use only pixels inside the binary ROI mask $m$ as input to the pixel classifier $G_t$ and then obtain the gradient of the cross-entropy loss $L_{ce}$ which was averaged by the number of pixels
% $N=n(A)$ for $A=\{(i,j) \mid m_{ij}=1\}$ 
$N=\sum_{ij} \mathbb{I} (m_{ij}=1)$
as follows:
%%------------------------------
\begin{equation}
    % \nabla_{seg} = \nabla_{x_t} \frac{1}{N} \sum_{\substack{1\le i\le H \\ 1\le j\le W}} L_{ce} (G_t(z_{ij}), y_{ij}) \times m_{ij}
    % \nabla_{seg} = \nabla_{x_t} \Biggl(\Bigl(\sum_{i=1}^{H} \sum_{j=1}^{W} L_{ce}(G_t(z_{ij}), y_{ij}) \times m_{ij} \Bigr)/ N \Biggr)
    L_{seg}(z, y^{edited}) = \frac{1}{N} \sum_{i=1}^{H} \sum_{j=1}^{W} m_{ij} L_{ce}\Bigl(G_t(z_{ij}), y_{ij}^{edited}\Bigr).
\label{eq:seg_loss}
\end{equation}
%%------------------------------

After obtaining the gradient, we guide the sampling process according to  Eq.\ref{eq:classifier_guidance} as follows.
%%------------------------------
\begin{equation}
    x_{\mathrm{fg}} \sim \mathcal{N}(\mu + s\Sigma \nabla_{x_t} L_{seg}, \Sigma),
\end{equation}
where $\mu = \mu_\theta(x_t, t), \Sigma = \Sigma_\theta(x_t, t)$.
%%------------------------------
This foreground term is related to the edit region.
Following \cite{avrahami2022blended}, we obtain background term by simply adding noise to the original image $x_0$ to preserve the outside of the edited region and combine these two terms to construct $x_{t-1}$ as follows:
%%------------------------------
\begin{align}
    x_{t-1,\mathrm{bg}} \sim \mathcal{N}(\sqrt{\bar{\alpha}_t}x_0, (1- \bar{\alpha}_t)I), \\
    x_{t-1} = x_{t-1,\mathrm{fg}} \odot m + x_{t-1,\mathrm{bg}} \odot (1-m),
\end{align}
%%------------------------------
where $\odot$ denotes the element-wise multiplication.
This guidance procedure is summarized in Algorithm.~\ref{algo:proposed_method}.

% \paragraph{Ranking system}
% Through above iterative guidance, we can obtain the multiple guided candidates with same intensity according to their batch-size.
% This is a desirable outcome thanks to the probabilistic sampling process of diffusion models.
% We sort these outcome according to the objective in eq.\ref{eq:seg_loss}, and defined a sample with the lowest $L_{ce}$ as a recommended outcome by the model.

\begin{algorithm}
\caption{Pixel-wise gradient-based guidance}
\begin{algorithmic}
\renewcommand{\algorithmicrequire}{\textbf{Input:}}
\renewcommand{\algorithmicensure}{\textbf{Output:}}
\REQUIRE original image $x$, target map $y^{edited}$, ROI mask $m$, guidance scale $s$, start denoising step $t_0$, pixel classifiers $G$
\ENSURE  edited image $x^{edited}$

% \STATE $N \leftarrow n(A), A=\{(i,j) \mid m_{ij}=1\}$
\STATE $N \leftarrow \sum_{ij} \mathbb{I} (m_{ij}=1)$
\STATE $x_0 \leftarrow x$
% \\ \textit{Obtain $x_{t_0}$}
% \\ \COMMENT{Obtaining latent variable}
\FOR {$t = 0$ to $t_0-1$}
\STATE $x_{t+1} \leftarrow \sqrt{\bar{\alpha}_{t+1}} f_\theta(x_t, t) + \sqrt{1-\bar{\alpha}_{t+1}} \epsilon_\theta (x_t, t)$
\ENDFOR
\FOR {$t = t_0$ down to $1$}
\STATE $\mu, \Sigma \leftarrow \mu_\theta(x_t, t), \Sigma_\theta(x_t, t)$
\STATE $L_{seg} \leftarrow\frac{1}{N} \sum_{i=1}^{H} \sum_{j=1}^{W} m_{ij} L_{ce}\Bigl(G_t(z_{ij}), y_{ij}^{edited}\Bigr) $
\STATE $x_{t-1,\mathrm{fg}} \sim \mathcal{N}(\mu + s\Sigma \nabla_{x_t} L_{seg}, \Sigma)$
\STATE $x_{t-1,\mathrm{bg}} \sim \mathcal{N}(\sqrt{\bar{\alpha}_t}x_0, (1- \bar{\alpha_t}) I)$
\STATE $x_{t-1} \leftarrow x_{t-1,\mathrm{fg}} \odot m + x_{t-1,\mathrm{bg}} \odot (1-m)$
\ENDFOR
\RETURN $x^{edited}$ 
\end{algorithmic}
\label{algo:proposed_method}
\end{algorithm}

%% Experiments
\section{Experiments}
\label{sec:experiments}
\begin{figure*}[t]
    \centering
    \includegraphics[width=0.94\linewidth]{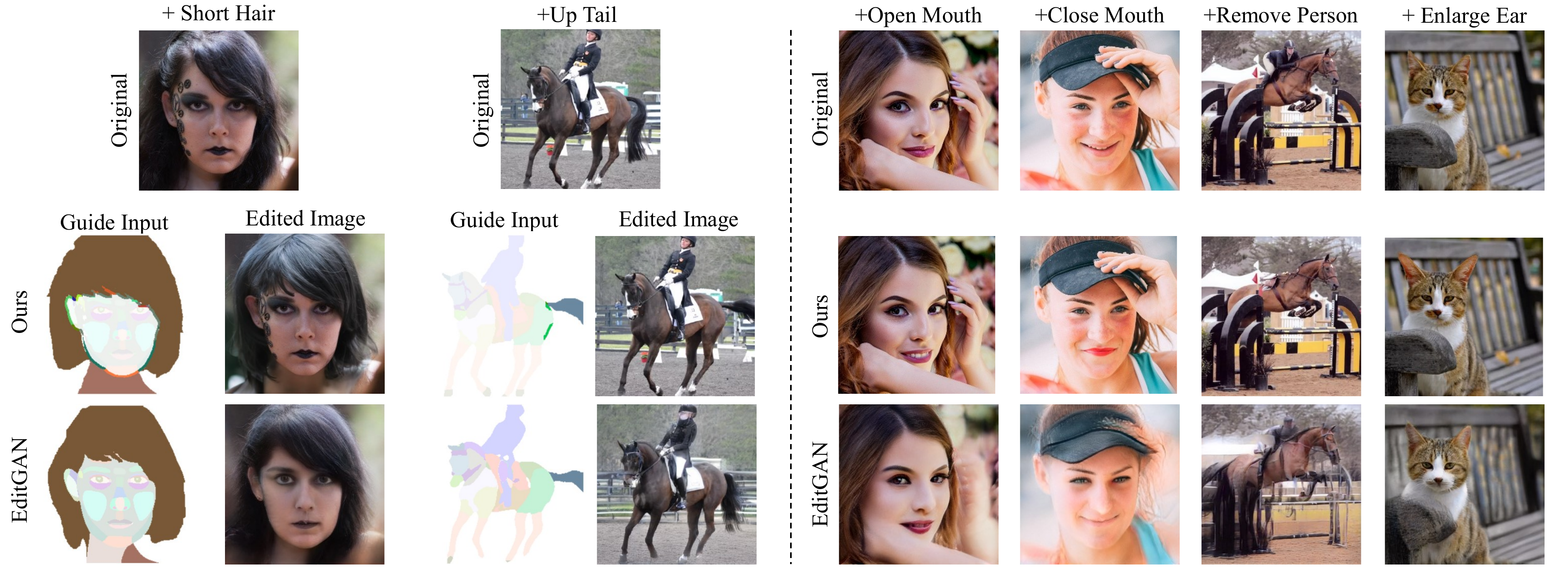}
    \caption{Qualitative comparions with EditGAN. These results were obtained by editing the segmentation map to follow the written text(\textit{e.g.} + Open Mouth).
    The left two samples show the edited map, which is included to clarify the specified manipulation. These results confirm that our method achieves the desired editing performance while preserving the details of the original images, whereas EditGAN fails to reconstruct some features, which are less observed in training data, such as facial painting, a person, and hands in original images.}
    \label{fig:experiments/comparison_w_editgan}
\end{figure*}

In this section, we demonstrate the quantitative and qualitative comparisons with some baseline methods.
We aim to achieve fine-grained editing in a label-efficient manner. 
Therefore, we chose EditGAN~\cite{Ling2021EditGANHS} and SDEdit~\cite{Meng2021SDEditIS} as the baseline methods, which can work with few ore zero annotated training data, respectively.

\subsection{Experimental Setting}
\paragraph{Datasets}
In all editing experiments, we used $256^2$ images.
We used FFHQ-256~\cite{Karras2019ASG}, LSUN-cat~\cite{Yu2015LSUNCO} and  LSUN-horse~\cite{Yu2015LSUNCO} for the EditGAN experiments, and  CelebA-HQ~\cite{Karras2018ProgressiveGO} for the SDEdit experiments.
For training of pixel-classifiers, 
we used the same number of annotated images and classes as DatasetDDPM~\cite{Baranchuk2022LabelEfficientSS}, 
but in the CelebA-HQ experiments, we used the backbone diffusion models and pixel-classifiers trained on FFHQ-256, 
because it is composed of more classes than the default CelebA-HQ number of classes of 19, and it provides higher editability.
% These numbers are listed in Table~\ref{tb:annotated_data}.
We used the annotated dataset as published in the DatasetDDPM official repository ~\footnote[3]{https://github.com/yandex-research/ddpm-segmentation}
%%------------------------

% \begin{table}[t]
% \centering
% \caption{Number of annotated images and classes for each dataset used in the training of the pixel-classifier and manipulation.}
% \renewcommand{\arraystretch}{1.2}
% \scalebox{0.8}{
% \begin{tabular}{c|c|c}
% \Hline
% Dataset    & Classes & Number \\ \hline
% FFHQ-256 ~\cite{Karras2019ASG}             & 34      & 20     \\
% LSUN-Horse ~\cite{Yu2015LSUNCO}            & 19      & 30     \\
% LSUN-Cat ~\cite{Yu2015LSUNCO}              & 15      & 30     \\
% CelebA-HQ ~\cite{Karras2018ProgressiveGO}  & 34      & - \\ 
% \Hline
% \end{tabular}}
% \label{tb:annotated_data}
% \end{table}

\vspace{-3mm}
\paragraph{Implementation}
To train the pixel classifiers, we trained only a single pixel classifier $G_t$ at each time step $t$ as a guidance model.
For guidance, we set hyper parameters \{$t_0$, $s$\} = \{$500$,$100$\} for the manipulation of small parts and \{$750$,$40$\} for large parts of FFHQ-256, CelebA-HQ and LSUN-Cat. 
And we show the editing performance using full-step guidance. For accelerated guidance, we use respaced generation~\cite{Dhariwal2021DiffusionMB} such that the total step is 50 steps.
The size of these parts was divided by a threshold of 5000 in terms of the number of pixels in the binary ROI mask $m$. 
On LSUN-Horse, we set $\{t_0, s\}$=$\{800,25\}$.
% because we mainly experimented manipulations that required large changes, such as generating objects in the background and vice versa on this dataset.
Additionally, in all experiments, we set the batch-size to four.
With these settings, the overall edit is finished in 15 $\sim$ 30 seconds on a single NVIDIA A100 GPU when we applied respaced editing.
% Sensitivity analysis on these parameters is discussed in Sec.\ref{subsec:experiments/dependency}

%% EditGAN
In the EditGAN experiments, we used optimization-based editing in \cite{Ling2021EditGANHS} because this approach, while increasing the editing time, offers the most accurate editing performance among the proposed approaches in EditGAN.
Then, we performed 100 steps of optimization using Adam~\cite{Kingma2015AdamAM} to optimize the latent codes.

In the SDEdit experiments, we used VP-SDE~\cite{Meng2021SDEditIS}, pre-trained on the CelebA-HQ dataset, for localized image manipulation on CelebA-HQ.
For all manipulations, we used stroke-based editing with $t_0 = 0.5$, $N=500$, and $K=1$.
In all manipulations, we create guide inputs as described in Fig.11 of ~\cite{Meng2021SDEditIS} with PaintApp.
The other implementation details are described in the appendix.

% use pretrained stylegan-2 ~\footnote{https://github.com/NVlabs/stylegan2} for backbone model as in their original papers. 
% But there are no official weight in FFHQ-256, so we use the reimplementation model\footnote{https://github.com/rosinality/stylegan2-pytorch} as a pre-trained model.
% For training, we train image encoder, for inversion, and DatasetGAN.
% As for image encoder training, we use same architecute~\cite{Richardson2021EncodingIS, Li2021SemanticSW} as in their papers.
% According to original papers, for cat and horse datasets, we train this encoder using only generated samples for first 20,000 iterations as warmup and then jointly use real images and generated images iteratively until model converges.
% As for DatasetGAN training, we use the same annotated data as in Table.\ref{tb:annotated_data}.
% Other training implementation details follow the original configurations.
% For inference, we use optimization-based editing in \cite{Ling2021EditGANHS} since this approach, while needing a editing time, offers the most accurate editing performance among proposed approaches in EditGAN.
% Then, we always perform 100 steps of optimization using Adam~\cite{Kingma2015AdamAM} to optimize the latent codes.

%%SDEdit

\vspace{-3mm}
\paragraph{Evaluation Metrics}
\label{subsubsec:evaluation metrics}
Our goal is to achieve fine-grained editing while preserving the content outside of the edited region and fast editing comparable to the GAN-based method.
Therefore, we evaluated the reconstruction performance, the accuracy of manipulation, and the quality of the edited whole images and runtime for quantitative evaluation.
We used MAE and PSNR to evaluate the reconstruction performance.
These metrics are calculated for pixels outside the editing region, defined by the binary ROI mask $m$ in Sec.\ref{subsubsec:map manipulation and mask preparation}.
For manipulation performance, we evaluated the prediction accuracy against the target map $y_{edited}$ by predicting the map corresponding to the edited images $x_{edited}$ with DatasetGAN and DatasetDDPM-R.
Subsequently, Inception Score  (IS)~\cite{Salimans2016ImprovedTF} and Fréchet Inception Distance (FID)~\cite{Heusel2017GANsTB} were used to ensure semantic consistency between the reconstructed and manipulated regions.
For the FID, we measured the distance between two distributions of the original images before editing and edited images.
We measured runtime under the same batch size set to 1.
% -----------------------------------------
\begin{figure*}[t]
    \centering
    \includegraphics[width=0.9\linewidth]{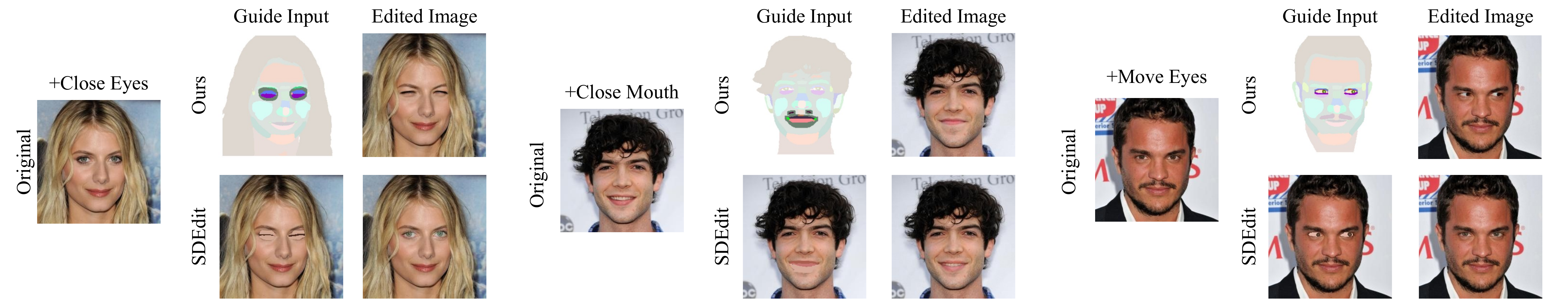}
    \caption{Qualitative comparison of our proposed method and SDEdit. We adopted stroke-based editing in SDEdit and experimented using VP-SDE ($t_0$=0.5, $N$=500). The guide input, which was edited with strokes using image editing software (\textit{e.g.} Paint app, Photoshop), were used for editing. As a result, we confirmed that SDEdit  failed to edit details (\textit{e.g.} move eyes) that our porposed method edited.}
    \label{fig:experiments/comparison_w_sdedit}
\end{figure*}
% -----------------------------------------
% --------------------------------
\begin{table*}[t]
\centering
\caption{Quantitative evaluation with EditGAN on FFHQ-256 for randomly selected 50 samples. In this table, we investigated the several results of our method depending on the guidance step and the strategy of selecting edited images within batch size. In (b), we also report the standard deviation. The results revealed that the proposed method outperforms EditGAN on all evaluation metrics.}
\renewcommand{\arraystretch}{1.4}
\scalebox{0.8}{
\begin{tabular}{llcclcclcl|c}
\Hline
\multicolumn{1}{c}{\multirow{2}{*}{Method}} & & \multicolumn{2}{c}{$m$=0} &  & \multicolumn{2}{c}{whole image} &  & \multicolumn{1}{c}{$m$=1} & & \multicolumn{1}{c}{\multirow{2}{*}{{Runtime [sec]}}}
\\ \cline{3-4} \cline{6-7} \cline{9-9} 
\multicolumn{1}{c}{} & & MAE (×$10^3$) ↓  & PSNR ↑  &  & FID ↓  & IS ↑  &  &  Accuracy ↑ & & \multicolumn{1}{c}{} \\ 
\midrule \midrule
EditGAN        &  & 74.85          & 67.24          &  & 79.67          & 3.765           &  & 81.38   & & 16.28      \\
\textbf{(a) Ours} (fullstep / quantitative) & & 14.25 & 81.44 &  & 15.81 & 4.377  &  & \textbf{83.02} &  & 146.48\\ 
\textbf{(b) Ours} (fullstep / random) & &  \textbf{13.93 }$\pm$ 1.742 & \textbf{81.59} $\pm$ 0.02958 &  & \textbf{14.62} $\pm$ 1.005 & 4.459 $\pm$ 0.03649 & & 81.95 $\pm$ 00.1742 &  & 146.48\\ 
% \textbf{Ours} (100 step) & &  0.01658 & \textbf{81.79} &  & 17.32 & 4.868  &  & 0.8070 & & 33.07 \\ 
\textbf{(c) Ours} (50 step / quantitative) & &  17.26 & 81.48 &  & 18.61 & \textbf{4.933}  &  & 80.74 & & \textbf{15.30} \\ 
\Hline
\end{tabular}}
\label{tb:quantitative evaluation}
\end{table*}

% (a) Ours (qualitative choice)                  & 0.01380                                                      & 81.53                                                     &                      & 15.07                                                   & 4.416                                                     &                     & 0.8290                                                               \\
% (b) Ours (quantitative choice)                 & 0.01425                                                      & 81.44                                                     &                               & 15.81                                                   & 4.377                                                     &                     & 0.8302                                                               \\
% (c) Ours (random choice)    & 0.01393 $\pm$ 0.001742 & 81.59 $\pm$ 0.02958 &  & 14.62 $\pm$ 1.005 & 4.459 $\pm$ 0.03649 & & 0.8195 $\pm$ 0.001742 \\ \Hline

\subsection{Evaluation and Comparison}
\paragraph{Quantitative Comparison}
\label{subsubsec:quantitative comparison}
% The results of the quantitative comparison with the EditGAN are presented in  Table~\ref{tb:quantitative evaluation}.
% The evaluation metrics are described in the Sec.\ref{subsubsec:evaluation metrics}.
In this experiment, we randomly selected 50 images on FFHQ-256 and then predicted the original segmentation map with segmentation models (\textit{i.e.} DatasetGAN and  DatasetDDPM-R), and applied one of the following manipulations: \{"close mouth," "open mouth," "move eye," "close eye," "edit eyebrow," "change hairstyle"\} on the segmentation map.
Our method can generate diverse outcomes owing to the probabilistic process of diffusion models, then we present our results in several settings in Table.~\ref{tb:quantitative evaluation}, such as (a) selecting a quantitatively superior sample within the batch size using full-step guidance, (b) random selection using full-step guidance and (c) selecting a quantitatively superior sample within the batch size using respaced generation.
As a result, we confirmed that the proposed method outperformed EditGAN in all metrics and it is not sensitive to the selection strategies.
Additionally, we achieved comparable editing speed to EditGAN~\cite{Ling2021EditGANHS} using respaced generation. This accelerated generation is useful for real-world applications.
% confirmed that the proposed method is not sensitive to the selection strategies and outperformed EditGAN in all metrics and accelearted generation can be achieved comparable editing speed to EditGAN~\cite{Ling2021EditGANHS}

% As in Table.~\ref{tb:quantitative evaluation}, the results revealed that, for all of these metrics, the proposed method outperformed EditGAN.
% Although the generation speed of our method is slower than EditGAN, it can generate diverse outcomes owing to the probabilistic process of diffusion models.
% Therefore, we selected a sample that is quantitatively superior within the batch size for evaluation.
% As in Table~\ref{tb:quantitative evaluation},
% the results revealed that, for all of these metrics, the proposed method outperformed EdiGAN.
% These results demonstrate that the proposed method achieves more fine-grained editing while preserving the outside of the edited region than EditGAN.

%%---------------------------------
% # EditGAN
% # Data Length:  50
% # Accuracy:  0.8137795432240105
% # MAE:  0.07485004875808954
% # PSNR:  tensor(67.2414)
% # ----------
% # Ours scale 25
% # Data Length:  50
% # Accuracy:  0.8047725003111263
% # MAE:  0.013257122039794922
% # PSNR:  tensor(82.0303)
% # ----------
% # Ours scale 40
% # Data Length:  50
% # Accuracy:  0.8203260203406357
% # MAE:  0.013992030760273338
% # PSNR:  tensor(81.1850)
% # ----------
% # Ours scale 25 / 40 Blending
% # Data Length:  50
% # Accuracy:  0.8289556773893616
% # MAE:  0.01379964116960764
% # PSNR:  tensor(81.5264)

%%---------------------------------

\paragraph{Qualitative Comparison}
First, we show the a comparison with EditGAN~\cite{Ling2021EditGANHS} in Fig.\ref{fig:overall_results} and Fig.\ref{fig:experiments/comparison_w_editgan}.
In these figures, we show the results obtained by the operation of the segmentation map as described in the text (\textit{e.g.}, Open Mouth).
From these figures, we confirmed that EditGAN failed to reconstruct some features, which are less observed in training data, such as the pacifier, face paintings, and the hands around the face, because of an inaccurate inversion performance.
This could be a significant challenge for applications that edit photos captured in the real world.
In addition, for some operations, such as editing a ponytail and short hair in Fig.\ref{fig:experiments/comparison_w_editgan}, EditGAN fails to achieve the desired editing on a default set of parameters.
In contrast, our method achieved the desired fine-grained editing naturally while preserving other features that were not reconstructed in EditGAN.

We then show a comparison with SDEdit~\cite{Meng2021SDEditIS} in Fig.\ref{fig:experiments/comparison_w_sdedit}.
For SDEdit, we use stroke-based edited images (Guide Input) as input to the SDE.
The manipulations are shown in Fig.\ref{fig:experiments/comparison_w_sdedit} and our aim is to compare the fine-grained editability.
The results reveal that SDEdit fails to edit such precise features of images, which can be achieved with our method.
%%%%%%%%%%%%%%%%%%%%% Trade Off (Appendix)

%%-------------------------
\begin{figure*}[t]
    \centering
    \includegraphics[width=0.9\linewidth]{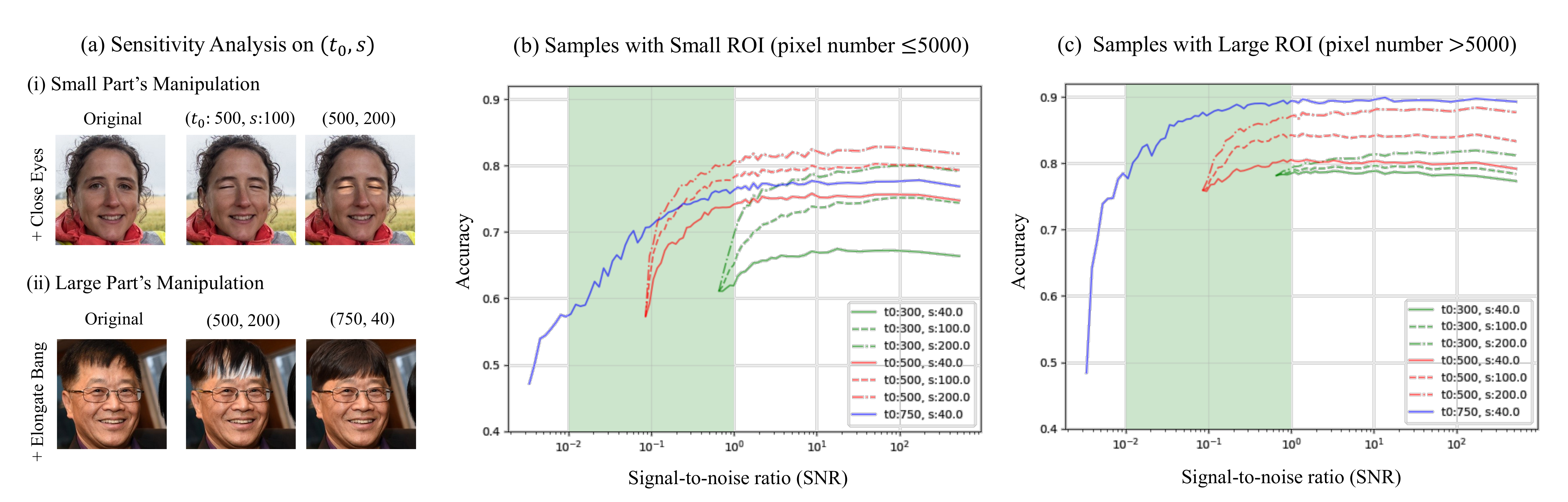}
    \caption{Analysis of the effect of the hyper-parameters, start denoising step $t_0$ and guidance scale $s$, on the prediction accuracy. This figure shows the relationship between signal-to-noise ratio (SNR) and accuracy on 50 experimental samples which were divided into two groups according to the size of editing region. This accuracy is averaged on each grouped samples and calculated against the estimated map on each step. When the SNR magnitude is between $10^{-2}$ and $10^0$, the perceptually recognizable content are generated~\cite{Choi2022PerceptionPT}.}
    \label{fig:experiments/hyperparameter}
\end{figure*}
%%-------------------------

\subsection{Further Applications}
\begin{figure}[t]
    \centering
    \includegraphics[width=0.97\linewidth]{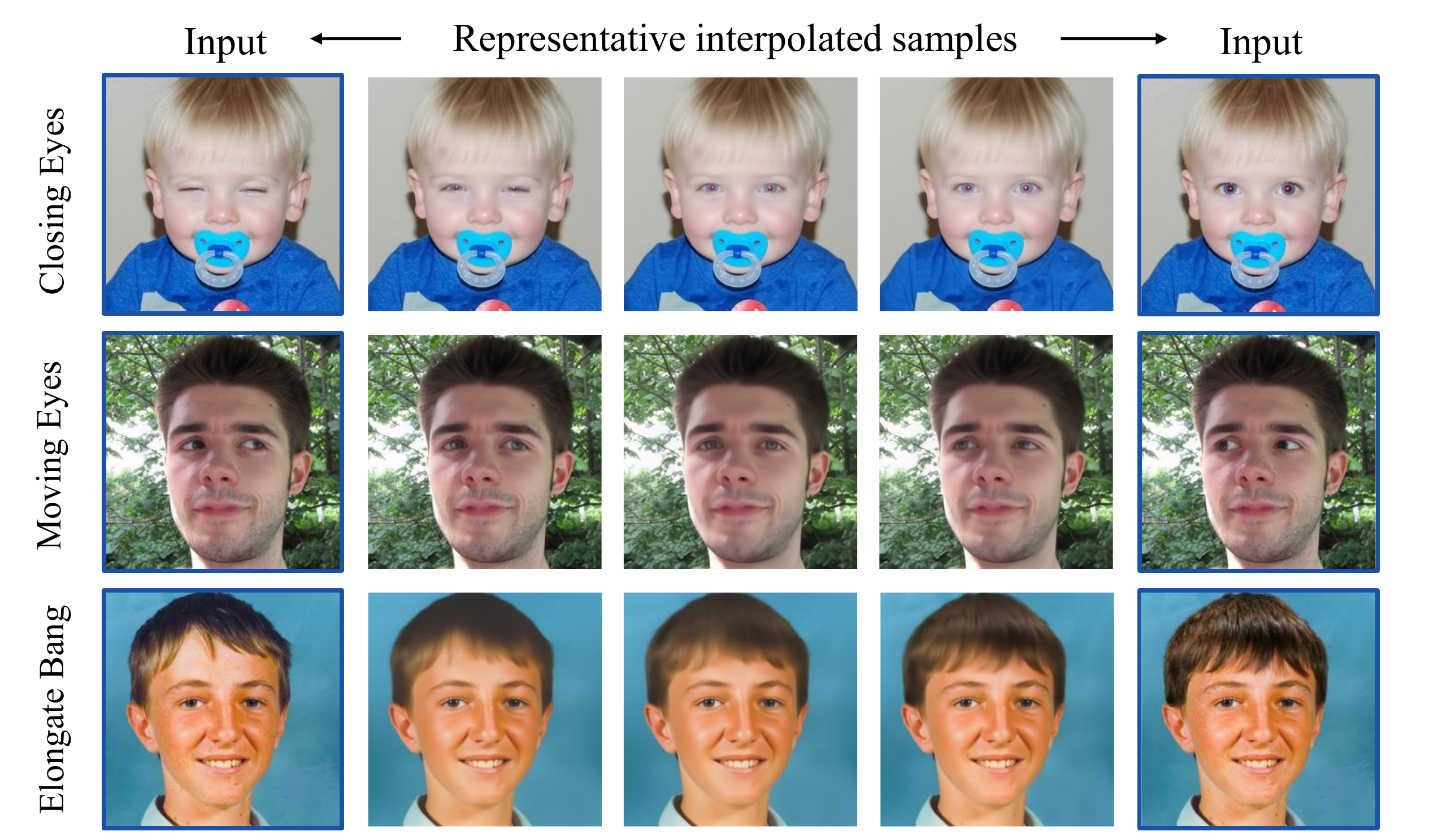}
    \caption{Results of interpolation between edited and original images. The results demonstrated the capability of meaningful interpolation on the latent space of pre-trained diffusion models.}
    \label{fig:experiments/interpolation_inpainting}
\end{figure}

The proposed methods can be utilized for some applications such as interpolation and inpainting.
For interpolation, we use real images and their manipulated results (or both manipulated results) as the two inputs.
We then interpolate the latent variables $x_{t_0}$ of these inputs, which are obtained using Eq.\ref{eq:ddim_forward}, to 50 samples, and reconstruct their images with Eq.\ref{eq:ddim_reverse}.
In Fig. \ref{fig:experiments/interpolation_inpainting}, we show the representative samples and interpolate variables at time step $t_0$ = 500.
These results demonstrate that we can obtain meaningful interpolated samples on the latent space of the diffusion model if these changes are minute. 
However, if there is a large difference between inputs, as shown in the examples of hairstyle change, the interpolated samples become slightly blurred.
This is because there is no guarantee that the manipulated samples will follow a linear relationship with the original images.
However, these editing vectors might be useful for controlling the intensity of editing, as in EditGAN.
For the inpainting, one of the results is presented in Fig.\ref{fig:experiments/comparison_w_editgan} ("remove person").
It requires a large start step of $t$=800, but we can confirm that the region with the person has been replaced naturally with the background.
%%----------------------
% \vspace{-5mm}
%%----------------------

\subsection{Detailed Analysis}
\label{subsec:experiments/dependency}
\paragraph{Sensitivity Analysis}
In this section, we analyze the sensitivity of hyper-parameters, the start denoising step $t_0$ and the guidance scale $s$, on the editing performance.
We demonstrate the relationship between signal-to-noise ratio (SNR) at each time step and prediction accuracy within the binary ROI mask $m$ in Fig.\ref{fig:experiments/hyperparameter}.
We experimented at $t_0$=\{300,500,750\} and $s$=\{40,100,200\}, respectively.
The larger $s$ than $s$=40 on $t_0$=750 has made artifacts because the large value of $s$ could guide the samples to the outside the learned distribution, so it is not shown in the figure.
In this experiment, we divided 50 experimental samples, which were the same samples in Sec.\ref{subsubsec:quantitative comparison}, into two groups: those with more than 5000 pixels within the binary ROI mask $m$ (Fig.~\ref{fig:experiments/hyperparameter}) and those with less than 5000 (Fig.~\ref{fig:experiments/hyperparameter}).
The accuracy between the target map $y_{edited}$ and the output of $G_t$ at each step, and averaged over the samples in the group.

First, we shall discuss $t_0$.
From Fig.\ref{fig:experiments/hyperparameter} (b) and (c), we confirmed that the manipulation of large regions requires a large $t_0$, while that of a small region can be achieved with a small $t_0$. 
This is because a  large content of images are created in the early step of the reverse diffusion process~\cite{Choi2022PerceptionPT, Baranchuk2022LabelEfficientSS, Kwon2022DiffusionMA}, which is visualized on a green background~\cite{Choi2022PerceptionPT}. A small content was created after that.
In Fig.\ref{fig:experiments/hyperparameter} (a), the results of the manipulation of a large region demonstrated that no matter how large the guidance scale $s$ is (such as $s$=200), it is difficult to change global features with a small step $t_0$=500, as this results in an  unnatural edit.

Next, we analyze the effect of the guidance scale $s$.
The results of manipulating a small region in Fig.\ref{fig:experiments/hyperparameter} (a) also reveal that a large scale $s$ generates images that are too aligned to a specified class and fail to generate realistic outcomes.
Therefore, we considered that $s$=100 was suitable on when we set $t_0$=500.
% When $t_0$ = 750, a value larger than 40 (such as $s$=100) would generate some artifacts because the large value of $s$ could guide the samples to the outside the learned distribution. 
We, therefore, set ($t_0$,$s$) = ($500$,$100$) for the manipulation of small regions and ($750$,$40$) for the manipulation of large regions on the face and cat datasets.
However, for LSUN horse manipulation, 
where an object is placed in the background and an object is generated on the background, we set a large value of $t_0$=800 and $s$ was set to 25 because a large value of $s$ is likely to shift the mean of $x_t$ outside the learned distribution on large $t_0$.

% In fig.\ref{fig:experiments/hyperparameter}, we compare the edited results as we guide the original image with shown guide input. 
% This figure ensures that the guidance from late reverse step ($t_0$ is smaller than 500) is unreasonable as we set the large value of guidance scale $s$ because the color of hair is unnaturally changed.
% On the other hand, the performance of pixel-classifiers would be degraded from the steps which is earlier than $t_0$ = 800 as described in \cite{Baranchuk2022LabelEfficientSS}.
% For the reasons above, we set $t_0$ =750 for guidance and $t_0$=800 for inpainting.

% Then look at the guidance scale $s$, the larger than $s$ = 25, we can obtain appropriate outcome in $t_0$ in fig.\ref{fig:experiments/hyperparameter}. 
% And $s$ = 40 can also achieve higher accuracy than $s$=25 in Table.\ref{tb:quantitative evaluation}.
% But we confirmed that some outcomes would be unrealistic one with $s$=50, $t_0$ = 750.
% So we set $s$ = 25 as a minimum required value which can achieve the desired edit.
% \vspace{-5mm}
\paragraph{Limitations}
There are a few cases that a large denoising step is necessary even if the manipulation is small when we manipulate the eyebrow or mouth with orthodontics.
In such cases, the user is required to manually adjust the step after observing the outcomes.
Additionally, as the proposed edit is based on class labels, the manipulation is restricted within the learned class label, and there are cases in which the original content color has been changed through editing. (\textit{\textit{e.g.,} eye color, hair color}).

% The main drawback of proposed method is the slow inference time.
% The running time varies depending on the size of the binary roi mask $m$,
% but it takes almost 1~2 minutes to manipulate one-sample.
% To obtain multiple candidates, we must take several minutes.
% It could be a major drawback on real-time editing, 
% but it would be resolved by future work.

%% Conclusion
\section{Conclusion}
\label{sec:conclusion}

In this paper, we summarized the requirements of real-world fine-grained image editing frameworks and proposed a novel image editing method with pixel-wise guidance using diffusion models. 
The effective combination of our label-efficient guidance and other techniques enabled highly controllable editing with preserving the outside of the edited area at a fast speed, meeting our requirements.
% The experimental results have demonstrated the advantage of our model qualitatively and quantitatively. 
While the automatic setting of hyper-parameters and color controls remain as our future work.

%abstract%
% Our goal is to develop a fine-grained real image-editing method suitable for real-world applications.
% In this paper, we first summarize four requirements for such methods and propose a novel image editing method with pixel-wise guidance using diffusion models that satisfies these requirements.
% Specifically, we first train pixel-classifiers with a few annotated data and then infer the segmentation map of a target image.
% Users then manipulate the map to instruct how the image will be edited.
% We utilize a pre-trained diffusion model to generate edited images aligned with the user's intention with pixel-wise guidance.
% The effective combination of proposed guidance and other techniques enables highly controllable editing with preserving the outside of the edited area, which results in meeting our requirements.
% The experimental results demonstrate that our proposal outperforms the GAN-based method for editing quality and speed.

% \input{appendix_ws}

\section*{Acknowledgement}
We would like to thank Dr. Satoshi Hara for useful advice. We also thank Tamaki Kojima for helpful comments.

%% REFERENCES
{\small
\bibliographystyle{ieee_fullname_}
\bibliography{refs.bib}
}

\appendix
\twocolumn[{%
\renewcommand\twocolumn[1][]{#1}%
\maketitle
\setcounter{figure}{7}
\begin{center}
    \centering
    \captionsetup{type=figure}
    \includegraphics[width=\linewidth]{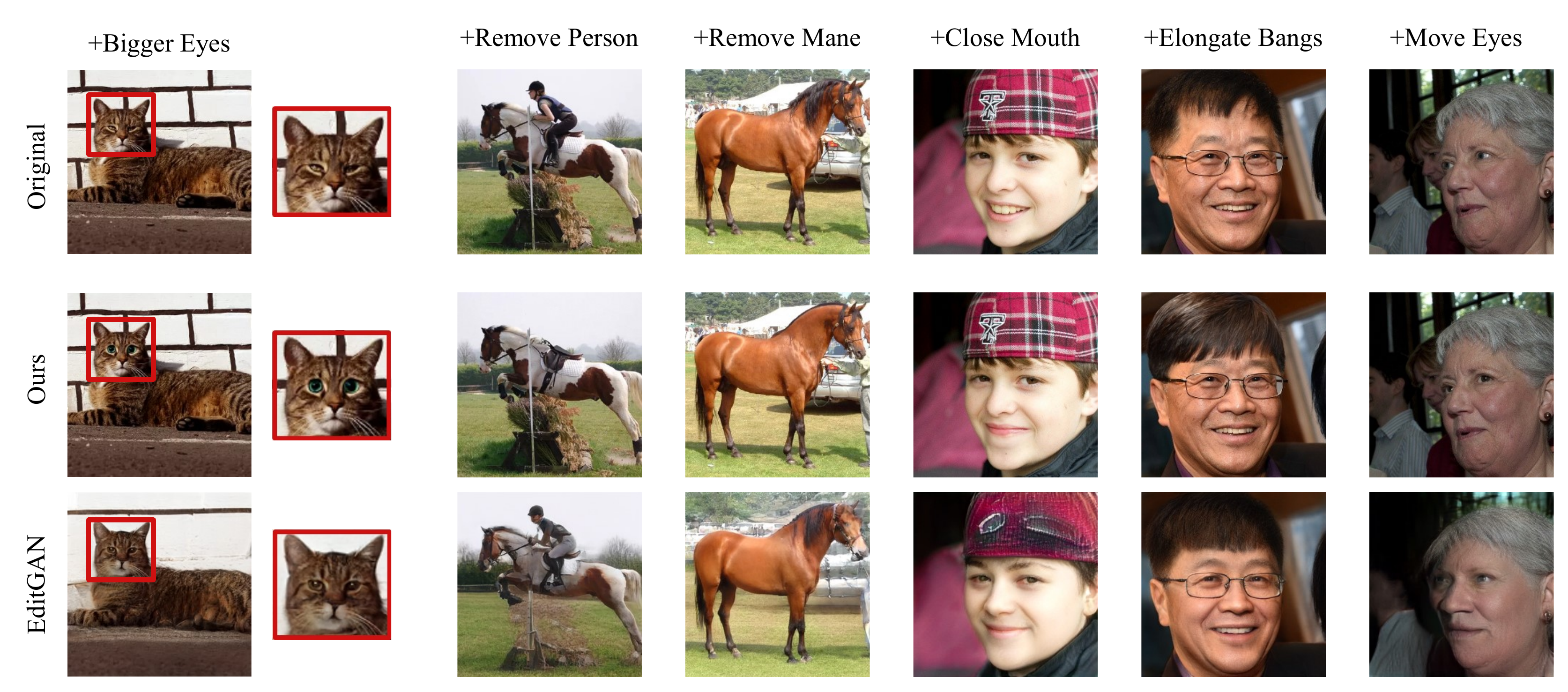}
    \captionof{figure}{Additional qualitative comparison with EditGAN.}
\label{fig:appendix/comparison_w_editgan}
\end{center}%
}]

%%-------------------

\begin{figure*}[t]
    \centering
    \includegraphics[width=\linewidth]{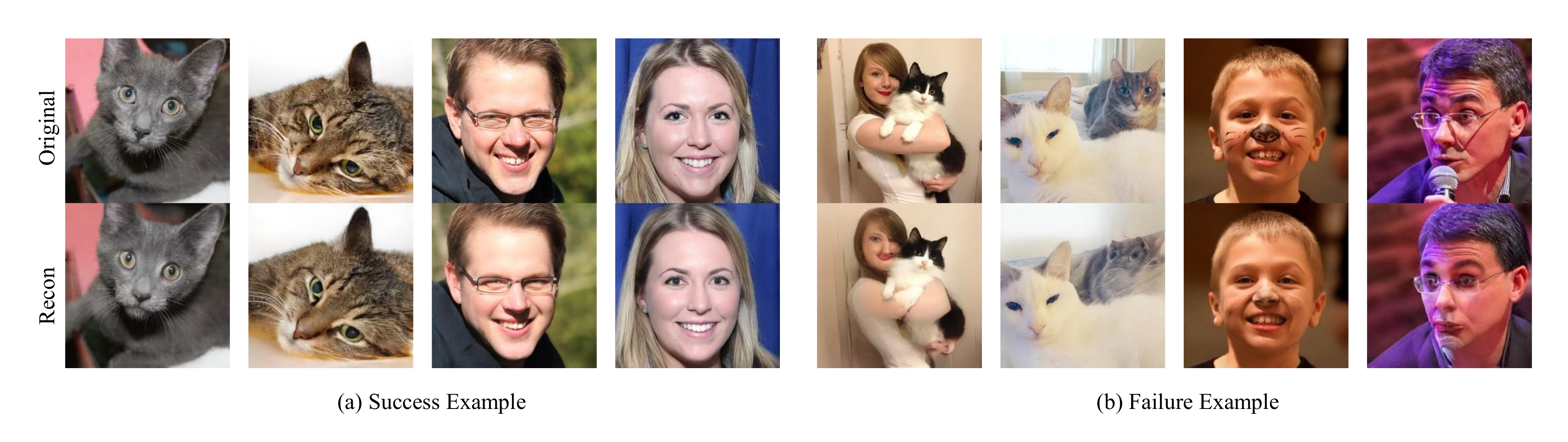}
    \caption{Results of the reconstruction performance in EditGAN. The latent codes corresponding to these original images are obtained by optimization after initialized by trained image encoder ~\cite{Ling2021EditGANHS}. These results demonstrate that the contents, which is less observed in the training data (\textit{e.g.,} "microrophone", "another cat"), were inaccurately reconstracted by EditGAN.
    }
    \label{fig:experiments/reconstruction}
\end{figure*}
%%-------------------
\begin{figure*}[t]
    \centering
    \includegraphics[width=\linewidth]{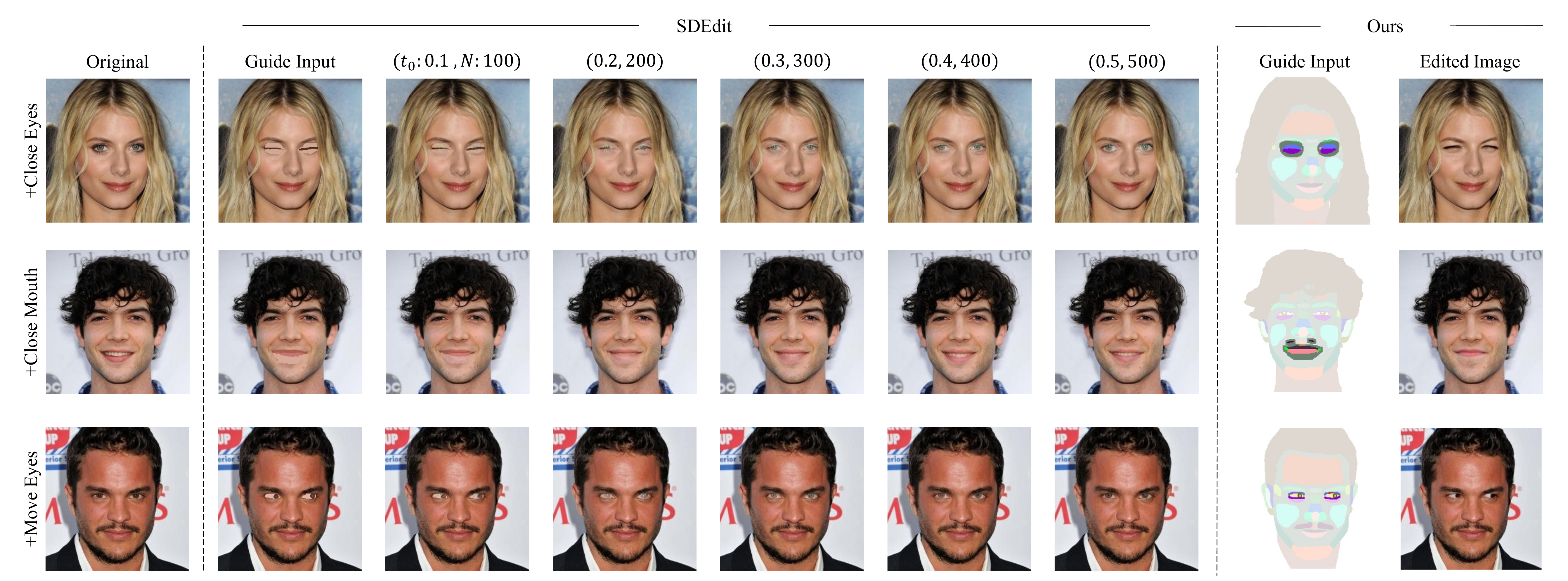}
    \caption{Comparison of stroke-based image editing performance with $t_0$ in SDEdit. We show the edited images with different values of $t_0$ in SDEdit. For SDEdit, we use VP-SDE ~\cite{Meng2021SDEditIS} and set $K$ = 1. These results reveal that SDEdit has a strict trade-off between the faithfulness and realism. It is difficult for the users to adjust the appropriate $t_0$ manually.}
    \label{fig:experiments/sdedit_t}
\end{figure*}
%%-------------------

\section{Implementation Details}
\subsection{Pixel Classifiers}
\label{subsec:appendix/implementation/pixel_classifier}
The proposed method uses ADM ~\cite{Dhariwal2021DiffusionMB} for backbone diffusion models and we used pre-trained ADM model provided in ~\cite{Baranchuk2022LabelEfficientSS}.
We respectively trained a single neural network for $g_t$ and $g_{multi}$, whereas the original DatasetDDPM~\cite{Baranchuk2022LabelEfficientSS} used an ensemble of ten individual networks. 
In this work, we used the intermediate representations extracted from the decoder block $B$ = {5,6,7,8,12} at each time step; therefore, the total dimensions of the pixel-wise features were 2816.
We used the same training conditions and model architecture as in their paper~\cite{Baranchuk2022LabelEfficientSS}.
We trained them for 4 epochs, using the Adam~\cite{Kingma2015AdamAM} optimizer with learning rate of 0.001. 
The batch size was set to 64. These settings are used for all datasets.

\subsection{EditGAN}
\label{subsec:appendix/implementation/editgan}
EditGAN~\cite{Ling2021EditGANHS} consists of three components, which are a generator of StyleGAN2, an image encoder for inversion, and DatasetGAN~\cite{Zhang2021DatasetGANEL} for estimation of segmentation maps.
As for the generator, we used the pretrained model, which is publicly available at the official repository \footnote[4]{https://github.com/NVlabs/stylegan2}.
Only for FFHQ-256, we used the pretrained model available at another repository, \footnote[5]{https://github.com/rosinality/stylegan2-pytorch} because this is not provided at the official one.

Regarding the image encoder, 
we used the same model architecture as in the literature ~\cite{Ling2021EditGANHS}.
We trained the encoder using the Adam optimizer with learning rate of $3 \times 10^{-5}$.
In their implementations, they train only on generated samples from GAN for first 20,000 iterations as warm up and then train jointly real and generated images on LSUN-Cat dataset. 
We applied this warm up training not only to LSUN-Cat but also to LSUN-Horse. 
In FFHQ, we trained without warm up.
For inference, we refine the latent code, which is initialized by this encoder, via 500-steps optimization as described in ~\cite{Ling2021EditGANHS}. 

As for pixel classifiers of DatasetGAN, we used the same annotated data for training. 
As in their setup~\cite{Ling2021EditGANHS}, we trained ten independent segmentation models, each of which is a three-layer MLP classifier ~\cite{Zhang2021DatasetGANEL}, using the Adam optimizer
with learning rate of 0.001.
Other training implementation details were same as the original configurations~\cite{Ling2021EditGANHS}.

\section{Additional Results}
\subsection{Reconstruction Performance with EditGAN}
\label{secsec:appendix/inversion}
Figure.~\ref{fig:experiments/reconstruction} shows example images on LSUN-Cat and FFHQ-256 that StyleGAN2 reconstructs based on the initialized latent codes by image encoder in EditGAN. 
Figure.\ref{fig:experiments/reconstruction} (a) and (b) show representative success and failure cases, respectively.
% We show the representative results which is successfully (Fig.~\ref{fig:experiments/reconstruction}(a)) and unsuccessfully reconstructed samples (Fig.~\ref{fig:experiments/reconstruction}(b)).
Although the reconstruction is accurate in the case of relatively simple scenes, it tends to fail, when the image contains objects that are rarely seen in the training dataset (\textit{e.g.,} face painting and microphone).
% Although EditGAN achieve accurate reconstruction on simple images in which the target object appears larger in the image, but some samples, whose features are less observed in training data such as person behind cat, microphone and face painting, are failed to reconstruct such features.
%%-------------------
\begin{figure*}[t]
    \centering
    \includegraphics[width=\linewidth]{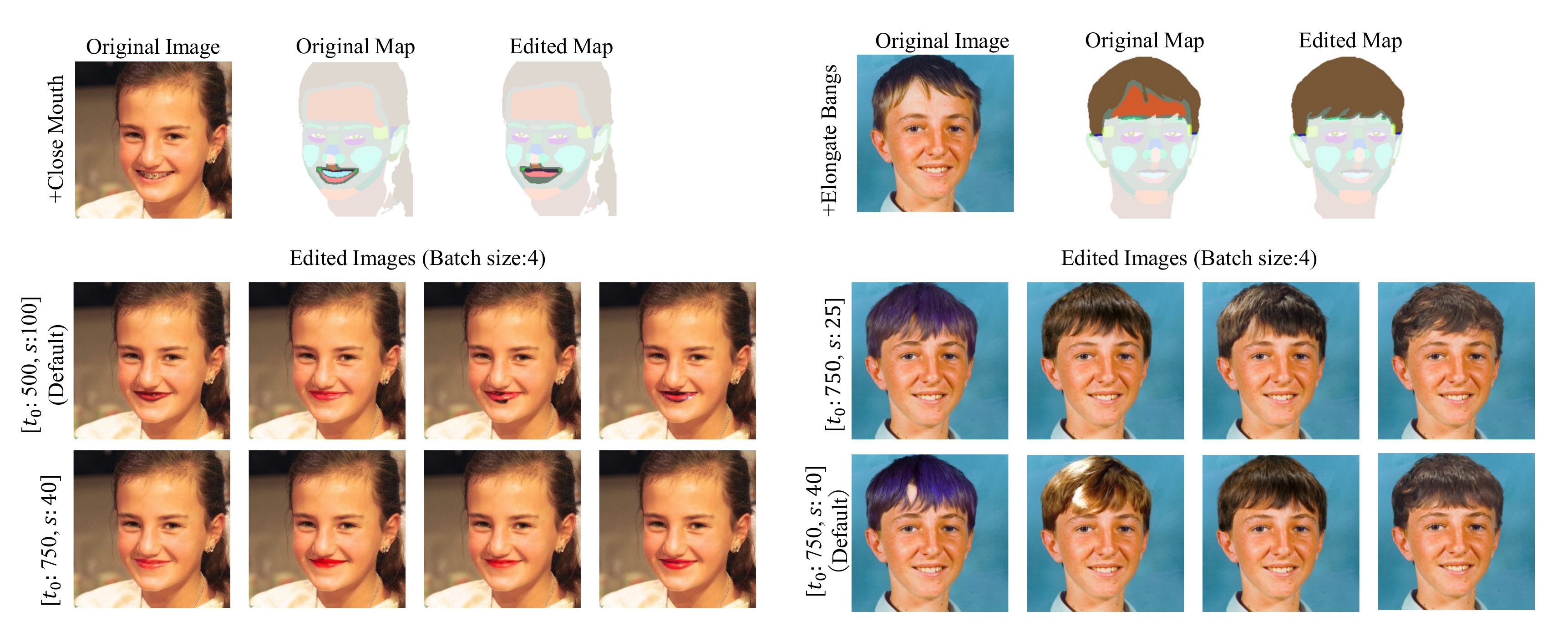}
    \caption{We show some failure cases. There are few exceptions which cannot be edited naturally on default set of $t_0$ and $s$. We show such edited examples obtained stochastically  within batch size, which we set to 4. Both examples can produce a natural image with some of the default settings. But, in the sample shown on the left, more steps are more stable, while in the sample shown on the right, smaller scales are more stable. In these cases, the user is required to manually tune the value of $t_0$ and $s$ after observing the outcomes.}
    \label{fig:experiments/failure}
\end{figure*}

%%------------------------------
\setcounter{table}{1}
\begin{table*}[t]
\centering
\caption{We present an analysis of performance changes for the selection strategy of edited images. We investigated several selection strategies in the proposed method. (a) selecting a qualitatively superior sample, (b) selecting a quantitatively superior sample, and (c) random selection. In (c), we also report the standard deviation of each performance metric. We confirmed that the proposed method is not sensitive to the selection strategies and outperformed EditGAN in all metrics.}
\renewcommand{\arraystretch}{1.6}
\scalebox{0.85}{
\begin{tabular}{lcclcclc}
\Hline
\multicolumn{1}{c}{\multirow{2}{*}{Method}} & \multicolumn{2}{c}{$m$=0} &  & \multicolumn{2}{c}{whole image} &  & \multicolumn{1}{c}{$m$=1} \\ \cline{2-3} \cline{5-6} \cline{8-8} 
\multicolumn{1}{c}{}  & MAE ↓  & PSNR ↑  &  & FID ↓  & IS ↑  &  &  Accuracy ↑ \\ 
\midrule \midrule
EditGAN                                     & 0.07485                                                      & 67.24                                                     &                               & 79.67                                                   & 3.765                                                     &                               & 0.8138                                                               \\
(a) Ours (qualitative choice)                  & 0.01380                                                      & 81.53                                                     &                      & 15.07                                                   & 4.416                                                     &                     & 0.8290                                                               \\
(b) Ours (quantitative choice)                 & 0.01425                                                      & 81.44                                                     &                               & 15.81                                                   & 4.377                                                     &                     & 0.8302                                                               \\
(c) Ours (random choice)    & 0.01393 $\pm$ 0.001742 & 81.59 $\pm$ 0.02958 &  & 14.62 $\pm$ 1.005 & 4.459 $\pm$ 0.03649 & & 0.8195 $\pm$ 0.001742 \\ \Hline
\end{tabular}}
\label{tb:quantitative evaluation_appendix}
\end{table*}
%%-------------------

\subsection{Further Comparison with SDEdit}
In SDEdit, there is a realism-faithfulness trade-off when we vary the value of $t_0$~\cite{Meng2021SDEditIS}.
Therefore, we also investigated how the edited images of our method change depending on the value of $t_0$.
In this experiment of SDEdit, 
we varied the value of $t_0$ from 0.1 to 0.5 and corresponding total denoising steps $N$ using VP-SDE ($K$=1).
Figure.\ref{fig:experiments/sdedit_t} shows several examples of the edited images. 
These results were obtained by selecting a qualitatively superior sample within a mini-batch for each image in both our method and SDEdit.
These results support that there is a strict trade-off between faithfulness and realism in SDEdit as reported in the paper~\cite{Meng2021SDEditIS}.
It is difficult for the users to tune $t_0$ manually because the appropriate $t_0$ varies from image to image. 
Also, even if the $t_0$ is set optimally, the proposed method can achieve more natural editing.
% Because there is no guarantee that the edited content on guide input will be preserved through a simple stochastic reverse SDE process as described in Sec.2.2.
% Consequently, we considered that it is difficult to achieve fine-grained image editing like these samples with SDEdit.

% \subsection{Running Time}
% We measured the averaged processing time of the proposed method over 50 samples, which are used for quantitative comparison in Sec.4.1.
% The averaged editing time was 147 seconds, when we set the batch size to 1. It increased to 341 seconds with larger batch size of 4.

\subsection{On the threshold to the size of the editing region}
\begin{figure}[t]
    \centering
    \includegraphics[width=\linewidth]{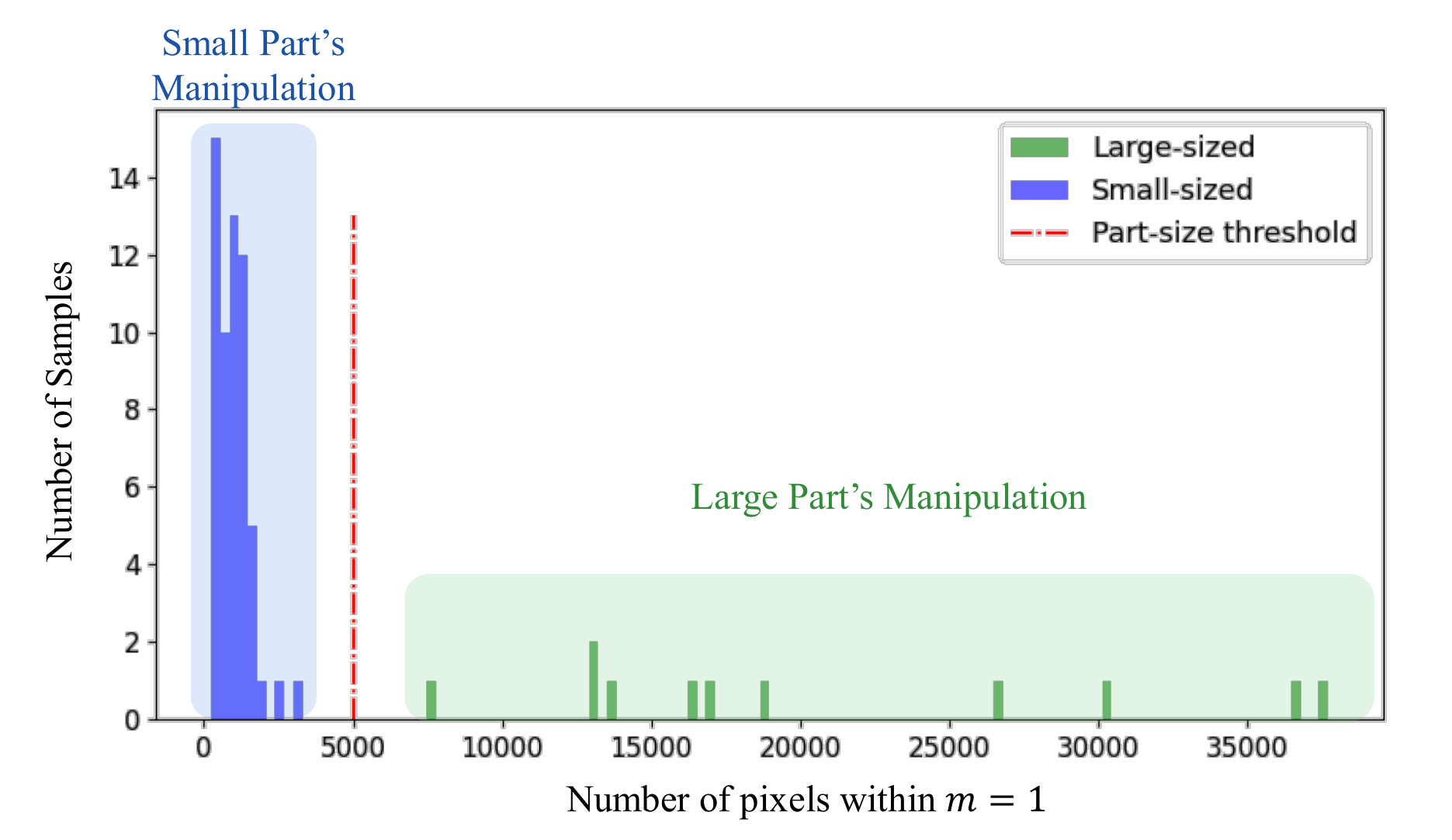}
    \caption{Statistics of the number of pixels in the ROI over all datasets used in Sec.4.2 and Sec.~\ref{subsec:appendix/qualitative}, which are FFHQ-34, Celeb-A, and LSUN-Cat. We confirmed that this threshold is not a sensitive parameter.}
    \label{fig:experiments/threshold}
\end{figure}

As described in Sec.4.4, we set the threshold to the size of the editing region and changed the value of $t_0$ based on this thresholding. 
% We divided experimental samples into two groups by setting a threshold for the number of pixels within the binary ROI mask $m$.
% Then we decided the start denoising step $t_0$ based on this threshold.
In the analysis in Sec.4.4, we set 5000 pixels as the threshold, because there was a big gap in ths size of the editing region between large and small part's manipulation.
In Fig.\ref{fig:experiments/threshold}, we show the statistics of the number of pixels in the ROI over all datasets used in the experiments of Sec. 4.2 and Sec.~\ref{subsec:appendix/qualitative}, which are FFHQ-34, Celeb-A, and LSUN-Cat.
It demonstrates that most of the samples have a fairly small number of pixels within $m$ = 1, while the others have exceptionally large.
To adopt the specific configuration to such samples with large ROI, we set the threshold to 5000. This value is not much sensitive to the performance of the proposed method as shown in Fig.~\ref{fig:experiments/threshold}.
%%-------------------
%%-------------------
\subsection{Failure Cases}
There were few failure cases, in which we cannot obtain natural results by the proposed method with the default setting of $t_0$ and $s$.
We show such examples in Fig.~\ref{fig:experiments/failure}.
In this figure, we demonstrate the all edited images within the mini-batch, which we set to 4. 
The original image, the original segmentation map, and the edited map are shown in the first row, 
and the second and third row show the edited images with the default setting and tuned setting, respectively.
From the samples shown on the left, we confirmed that images with features less observed in training data, such as teeth with orthodontics, require larger $t_0$, such as $t_0$=750, to produce naturally edited images.
From the right samples, we found that setting a large value to $s$ occasionally changed hair color in some cases.
In these cases, the user is required to manually tune the value of $t_0$ and $s$ after observing the outcomes.

%%------------------------------

\subsection{On strategies for selecting edited images}
In this work, we assume that the users selected the qualitatively best images among the several edited images provided by our method.
Consequently, the performance of our method may depend on this selection strategy, because our method provides some variety in the edited images as shown in Fig.~\ref{fig:experiments/failure}.
We investigated the performance of our method with several strategies for the selection: 
(a) selecting a qualitatively superior sample as in Sec.4.2,
(b) selecting a quantitatively superior sample in terms of the error of the segmentation map estimated from the edited images,
(c) random selection.
The results are shown in Table~\ref{tb:quantitative evaluation_appendix}.
In (c) we also report the standard deviation of each performance metric.
We confirmed that the performance of the proposed method is not sensitive to the choice of the strategy.

% Table~\ref{tb:quantitative evaluation} shows the further quantitative comparison with EditGAN when we change the way of choosing a sample within mini-batch in the proposed method.
% We demonstrate the three patterns: (a) choosing a qualitatively superior sample as in Sec.4.2, (b) choosing a quantitatively superior sample based on the accuracy for the target map $y_{edited}$, and (c) choosing randomly.
% In (c), we calculated the mean and standard deviation within mini-batch.
% The overall results demonstrate that the proposed method is robust to the way of choosing the outcome within mini-batch and all patterns of the proposed method outperformed EditGAN in all metrics.

% B.6. On strategies for selecting edited images
% In this work, we assume that users select the qualitatively best resultant image among the several edited images provided by our method. Consequently, the performance of our method may depend on this selection, because our method provides some variety the in edited images as shown in Fig. 11. We investigated the performance of our method with several strategies for the selection: (a) selecting a qualitatively superior sample as in Sec. 4.2, (b) selecting a quantitatively superior sample in terms of the error of the segmentation map estimated from the edited images, and (c) random selection. The results are shown in Table 1. In (c), we also report the standard deviation of each performance metric. We confirmed that the performance of our method is robust against the choice of the strategy.

%%------------------------------
\subsection{Additional Results of Our Method}
\label{subsec:appendix/qualitative}
We present additional examples of the edited images produced by our method on FFHQ-256, LSUN-Cat and LSUN-Horse (Figs.13-19).
In all examples, the original real images and the corresponding maps are shown on the left, and the edited maps and the resultant images are shown on the right.
The experimental settings are the same with those in Sec.4.2 and Sec.4.3
% As in Sec.4.2, we selected a sample that is qualitatively superior within the four batch size.
% According to the Sec.4.3, we set \{$t_0$,$s$\} = \{$500$,$100$\} for the manipulation of small regions and \{$750$,$40$\} for the manipulation of large regions on the FFHQ-256 and LSUN-Cat datasets.
% These regions are divided by the ROI size threshold, which we set to 5000 pixels.
% On LSUN-Horse, most samples follow this threshold setting, but for the manipulation, which forces to replace some objects to background or generate some objects on background, (\textit{i.e.} "+Raise Tail",  "+Remove Person"), we set $\{t_0, s\}$=$\{800,25\}$, because these manipulation requires a large step as discussed in Sec.4.4.

%%-------------------
\clearpage
\begin{figure*}[t]
    \centering
    \includegraphics[width=\linewidth]{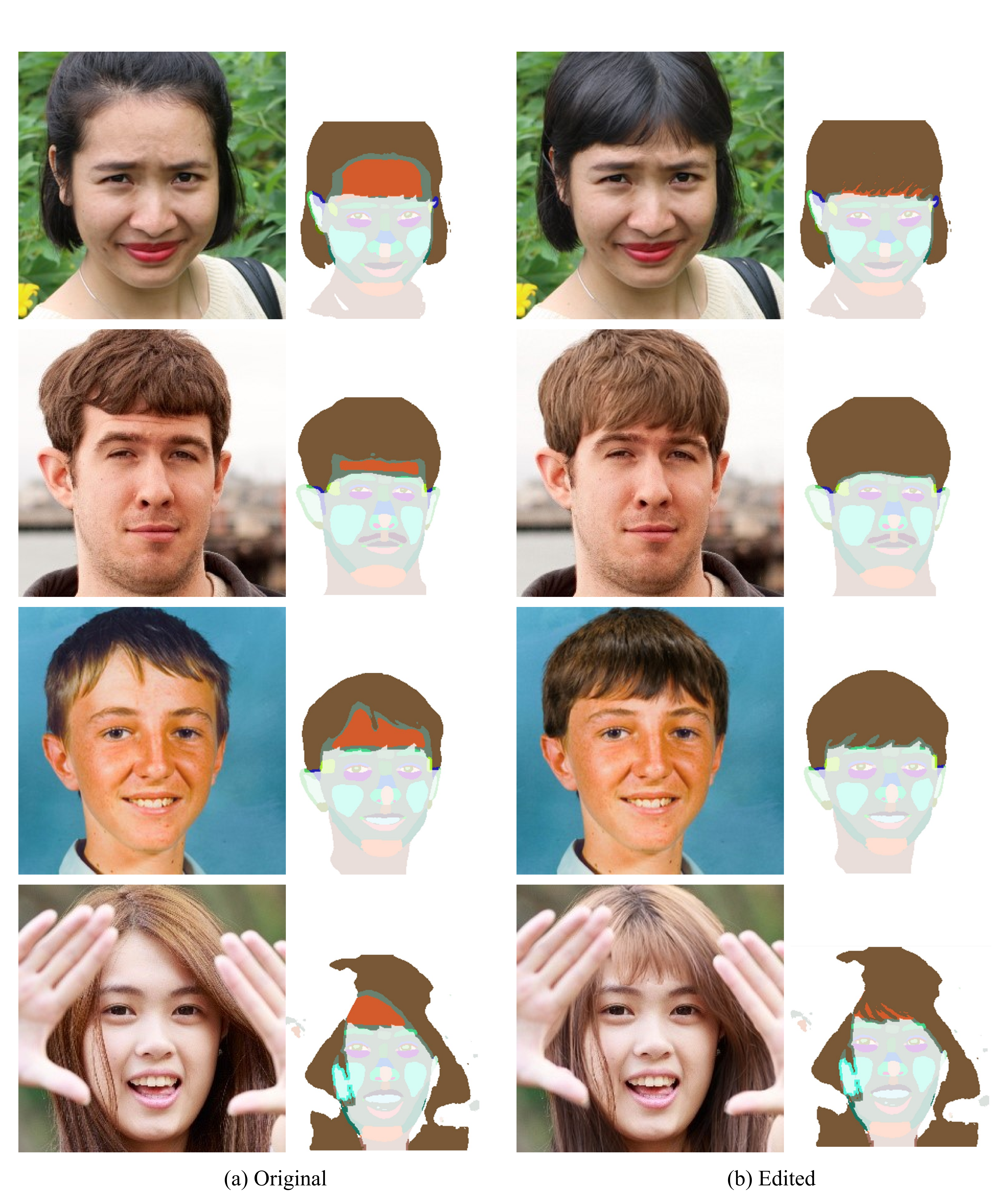}
    \caption{\textbf{Hairstyle Editing.} }
    \label{fig:experiments/hairstyle}
\end{figure*}
%%-------------------
\clearpage
\begin{figure*}[t]
    \centering
    \includegraphics[width=\linewidth]{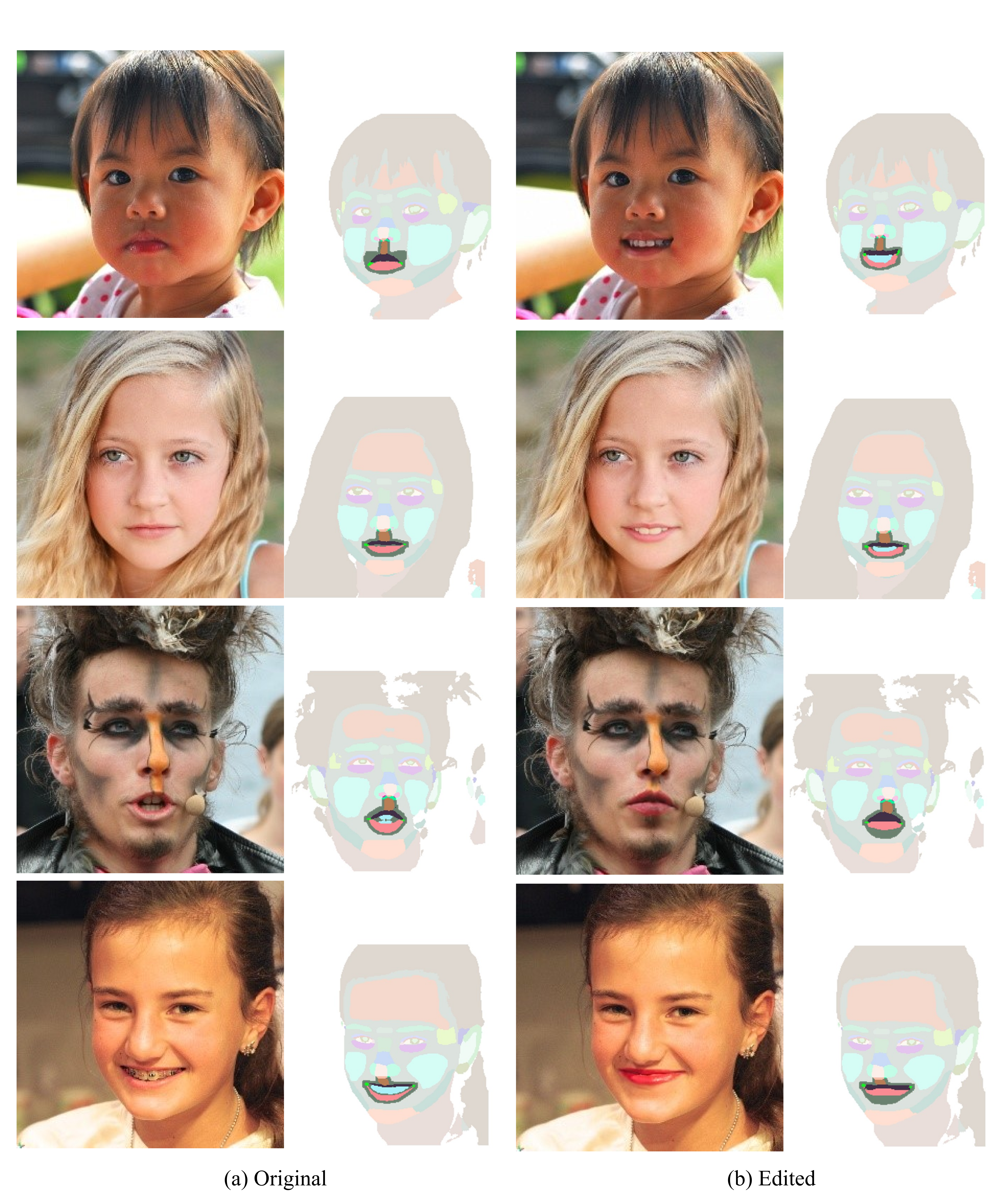}
    \caption{\textbf{Mouth Editing.} In these examples, we show the manipulation of "+Open Mouth" in the top two rows and examples of "+Open Mouth" in the bottom two rows.}
    \label{fig:experiments/mouth}
\end{figure*}
%%-------------------
\clearpage
\begin{figure*}[t]
    \centering
    \includegraphics[width=\linewidth]{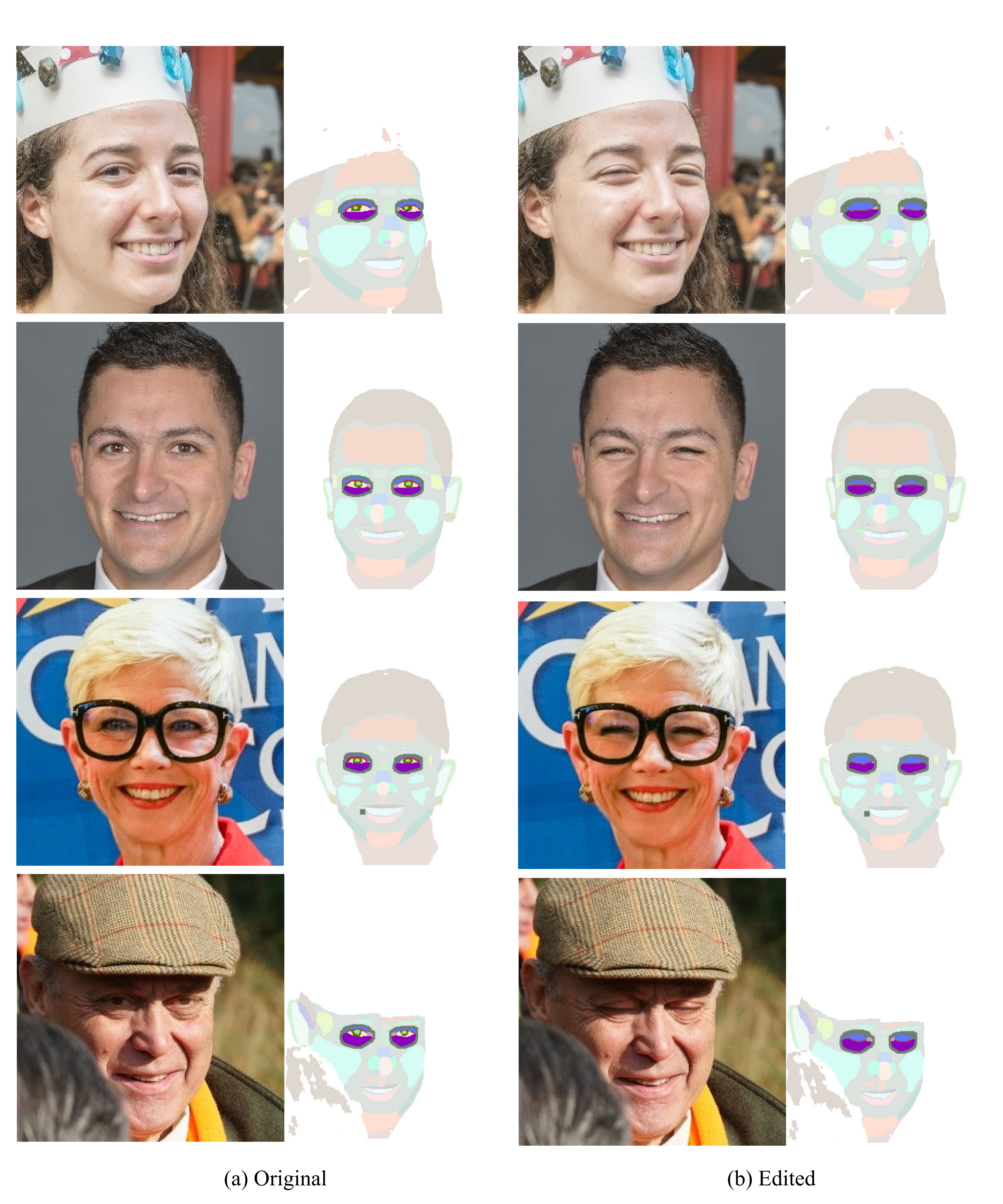}
    \caption{\textbf{Closing Eyes.} We show some examples of the manipulation of "+Close Eyes". We confirmed that eyes in glasses, eyes with shadows, etc., could also be generated naturally.}
    \label{fig:experiments/eyeclose}
\end{figure*}
%%-------------------
\clearpage
\begin{figure*}[t]
    \centering
    \includegraphics[width=\linewidth]{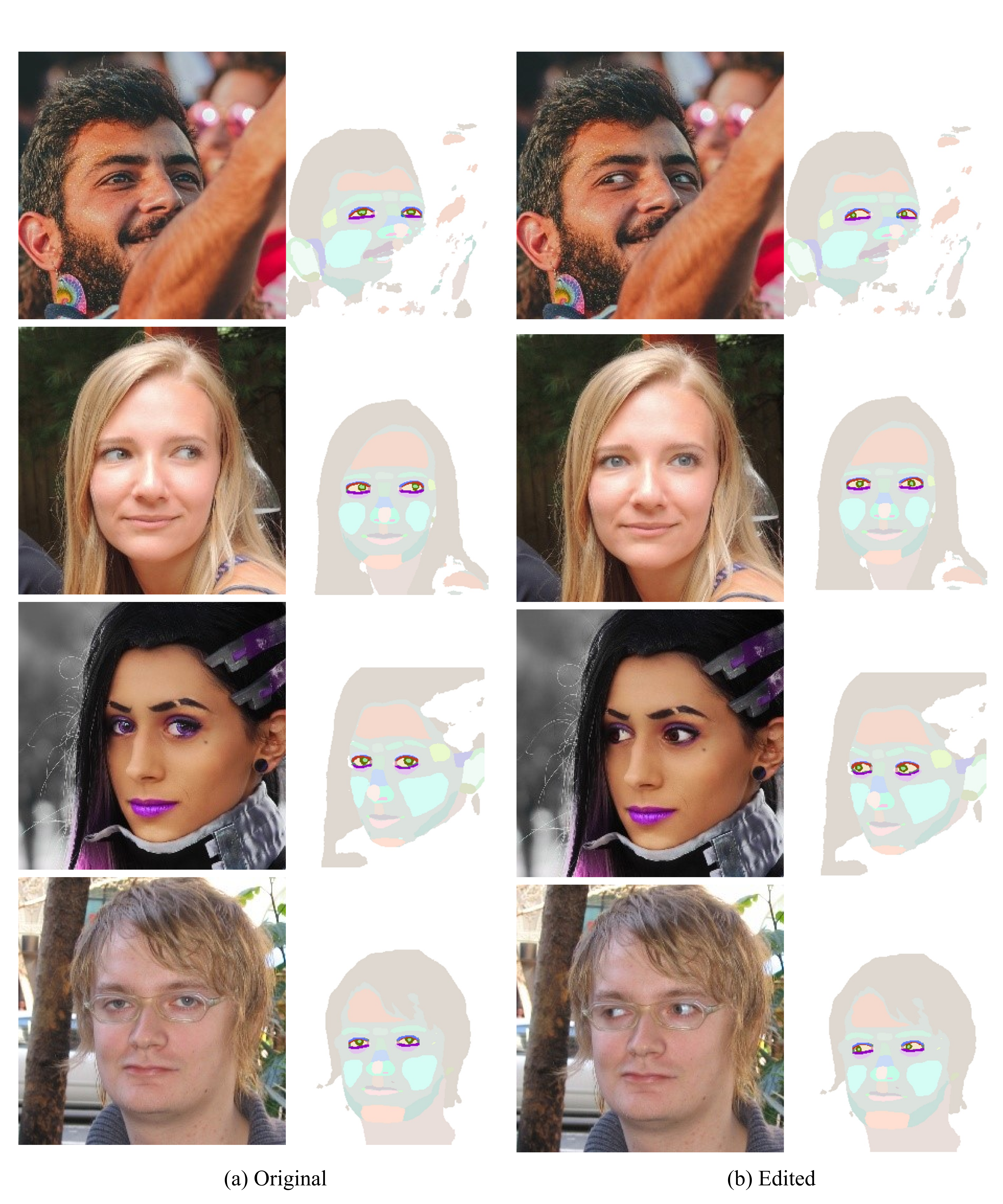}
    \caption{\textbf{Moving Eyes.} We demonstrate some examples corresponding to "+Moving Eyes".}
    \label{fig:experiments/eyemove}
\end{figure*}
%%-------------------
\clearpage
\begin{figure*}[t]
    \centering
    \includegraphics[width=\linewidth]{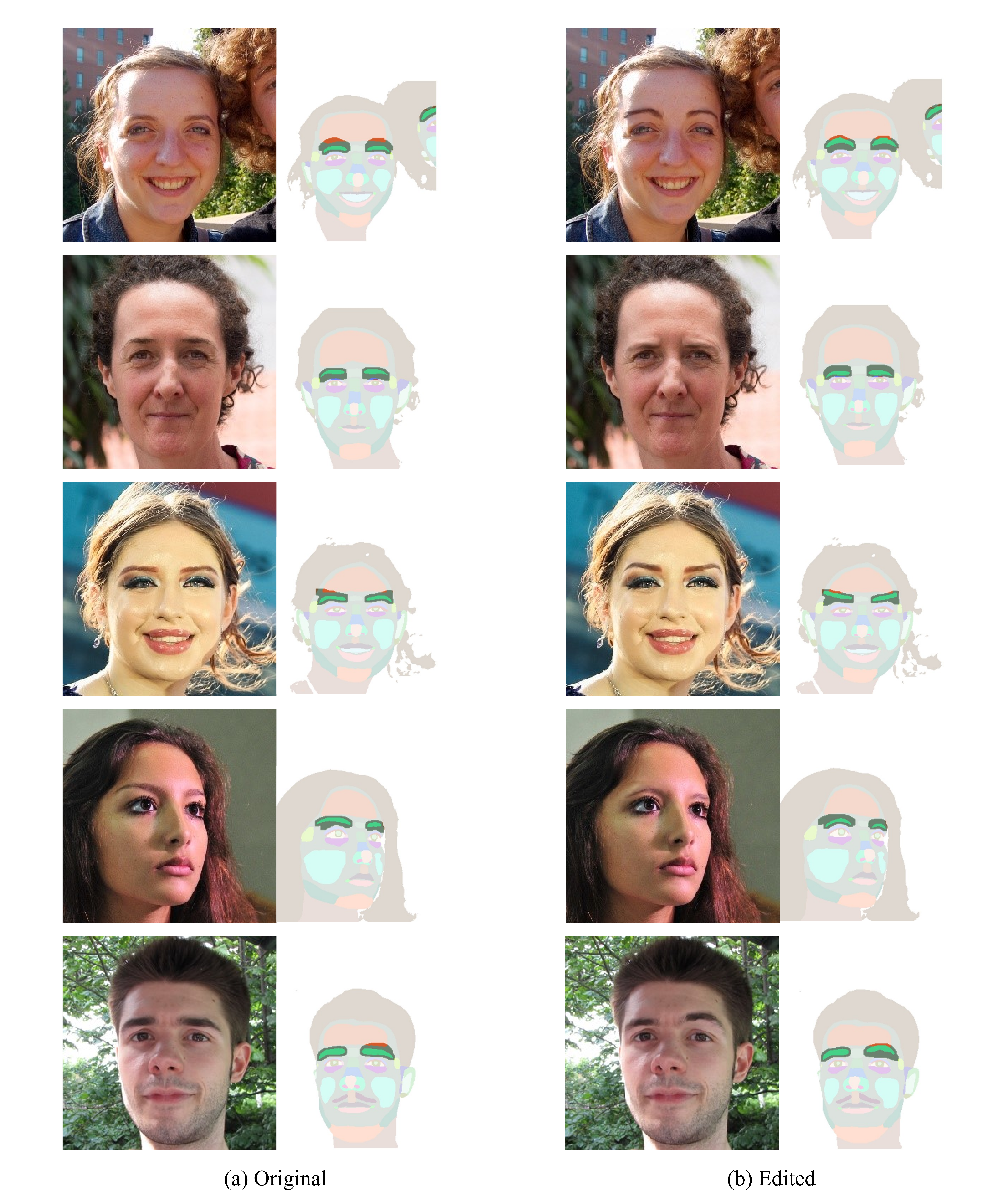}
    \caption{\textbf{Eyebrow Editing.} We show examples of editing that change the shape or position of eyebrows. We confirmed that the proposed method allows editing only one of the semantic-related parts (such as lift right eyebrow while keeping the left eyebrow unchanged), which EditGAN is not good at.}
    \label{fig:experiments/eyebrow}
\end{figure*}
%%-------------------
\clearpage
\begin{figure*}[t]
    \centering
    \includegraphics[width=0.9\linewidth]{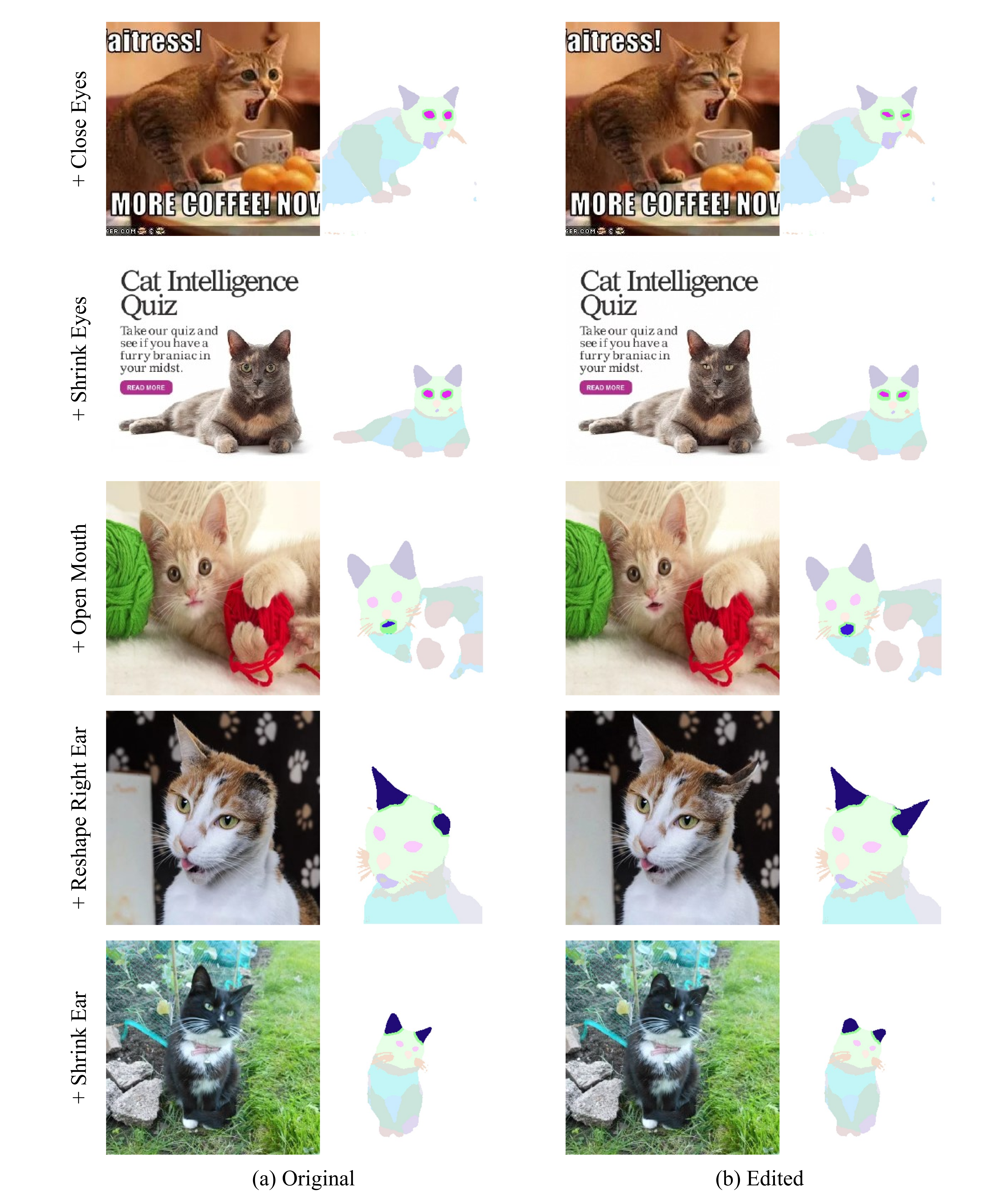}
    \caption{\textbf{Editing cat images.} We show some edited examples following by the written text.}
    \label{fig:experiments/cat}
\end{figure*}
%%-------------------
%%-------------------
\clearpage
\begin{figure*}[t]
    \centering
    \includegraphics[width=0.9\linewidth]{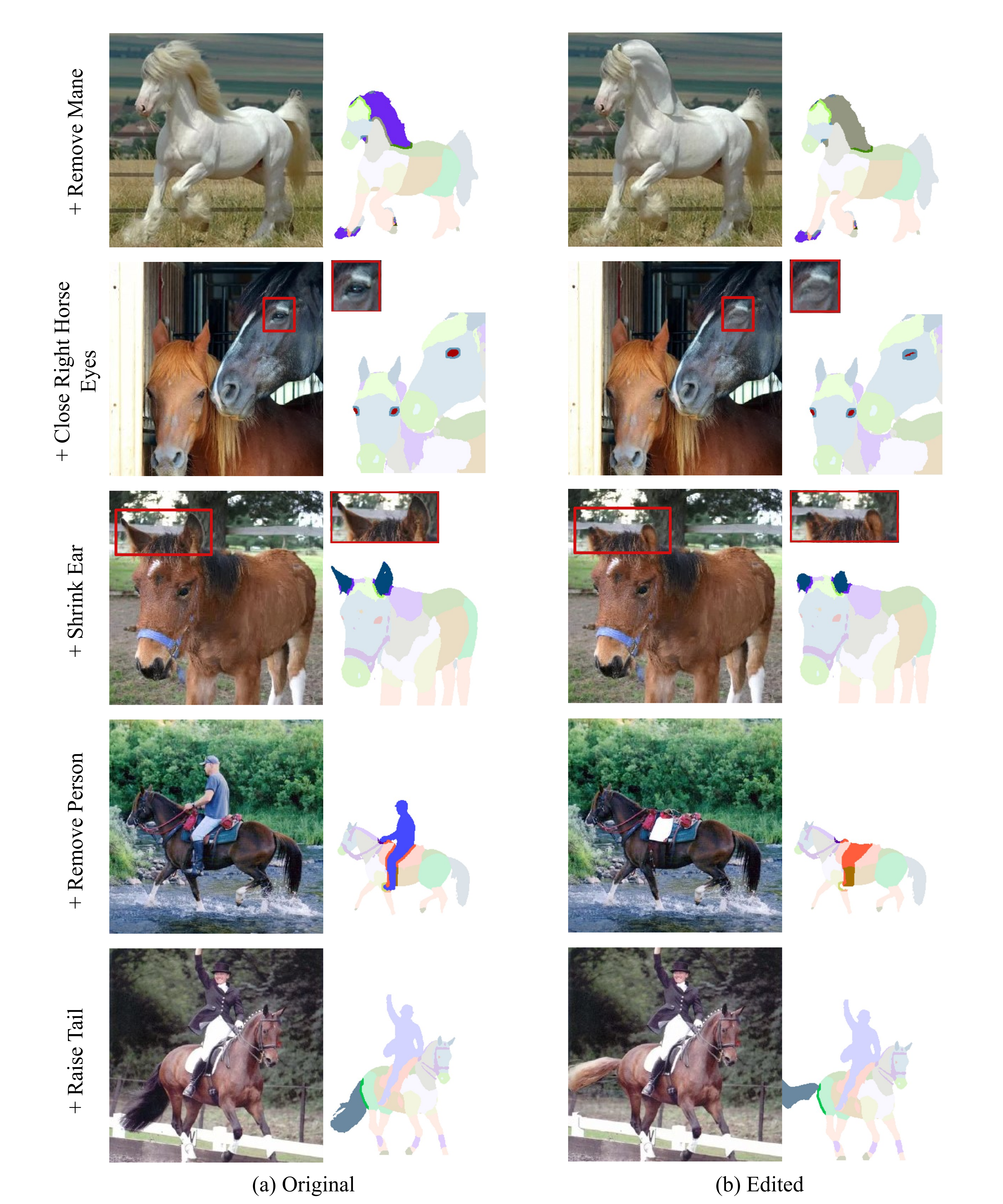}
    \caption{\textbf{Editing horse images.} We show some edited examples following by the written text. In the above three samples, we set \{$t_0$,$s$\} = \{$500$,$100$\} according to the threshold setting. In the two samples below, we set $\{t_0, s\}$=$\{800,25\}$, because this manipulation, which forces to replace some objects to background or generate some objects on background, requires a large steps as discussed in Sec.4.4. In the bottom example, the semantic editing was done, but the color was changed as described in Sec.\ref{fig:experiments/failure}}
    \label{fig:experiments/horse}
\end{figure*}
%%-------------------
\clearpage
% {\small
% \bibliographystyle{ieee_fullname_}
% \bibliography{refs.bib}
% }

\end{document}